\documentclass[letterpaper]{article} 
\usepackage{aaai25}  
\usepackage{times}  
\usepackage{helvet}  
\usepackage{courier}  
\usepackage[hyphens]{url}  
\usepackage{graphicx} 
\usepackage{natbib}  
\usepackage{caption} 
\usepackage{algorithm}
\usepackage{algorithmic}

\usepackage{booktabs}
\usepackage{multirow}
\usepackage{svg}
\usepackage{amsmath,amssymb}
\usepackage{here}
\usepackage{url}
\usepackage{comment}
\usepackage{multirow}
\usepackage{tabularx}
\usepackage{color}
\usepackage{xspace}
\usepackage[most]{tcolorbox}

\usepackage{newfloat}
\usepackage{listings}
\DeclareCaptionStyle{ruled}{labelfont=normalfont,labelsep=colon,strut=off} 
\floatstyle{ruled}
\newfloat{listing}{tb}{lst}{}
\floatname{listing}{Listing}
\title{ToMATO: Verbalizing the Mental States of Role-Playing LLMs\\ for Benchmarking Theory of Mind}
\author{
    Kazutoshi Shinoda, Nobukatsu Hojo, Kyosuke Nishida, Saki Mizuno, \\
    Keita Suzuki, Ryo Masumura, Hiroaki Sugiyama, Kuniko Saito
}
\affiliations{
    NTT Corporation\\

    kazutoshi.shinoda@ntt.com
}

\begin{document}
\maketitle

\begin{abstract}
Existing Theory of Mind (ToM) benchmarks diverge from real-world scenarios in three aspects: 1) they assess a limited range of mental states such as beliefs, 2) false beliefs are not comprehensively explored, and 3) the diverse personality traits of characters are overlooked.
To address these challenges, we introduce ToMATO, a new ToM benchmark formulated as multiple-choice QA over conversations.
ToMATO is generated via LLM-LLM conversations featuring information asymmetry.
By employing a prompting method that requires role-playing LLMs to verbalize their thoughts before each utterance, we capture both first- and second-order mental states across five categories: belief, intention, desire, emotion, and knowledge.
These verbalized thoughts serve as answers to questions designed to assess the mental states of characters within conversations.
Furthermore, the information asymmetry introduced by hiding thoughts from others induces the generation of false beliefs about various mental states.
Assigning distinct personality traits to LLMs further diversifies both utterances and thoughts.
ToMATO consists of 5.4k questions, 753 conversations, and 15 personality trait patterns.
Our analysis shows that this dataset construction approach frequently generates false beliefs due to the information asymmetry between role-playing LLMs, and effectively reflects diverse personalities.
We evaluate nine LLMs on ToMATO and find that even GPT-4o mini lags behind human performance, especially in understanding false beliefs, and lacks robustness to various personality traits.
\end{abstract}

\section{Introduction}
Theory of Mind (ToM) is the cognitive ability to infer unobservable mental states such as beliefs, intentions, and desires of others \cite{premack_woodruff_1978}.
The ToM reasoning capability is thought to be the cornerstone of human social intelligence \cite{fan-etal-2022-artificial}, and indispensable to interact with others \cite{baron1985autistic}.

To investigate whether large language models (LLMs) possess human-like ToM, researchers have used various benchmarks.
However, existing ToM benchmarks are not aligned well with real-world scenarios in the following three aspects.
(1) Despite various categories of mental states that can be inferred by ToM as studied in psychology \cite{beaudoin2020systematic}, only limited types of mental states such as beliefs have been assessed \cite{ma-etal-2023-towards-holistic}, especially for second-order ToM.
(2) False beliefs about beliefs or world states have been the main focus of previous studies \cite{le-etal-2019-revisiting,kim-etal-2023-fantom}.
However, false beliefs about other types of mental states have not been explored.
Understanding false beliefs about a range of mental states should be crucial for LLMs to facilitate effective social interaction in real-world scenarios.
(3) The behaviors and mental states of the characters in most benchmarks do not depend on their personality traits, even though they do in the real world \cite{costa1980influence,izard1993stability,mehl2006personality}.

\begin{figure*}
    \centering
    \includegraphics[width=\linewidth]{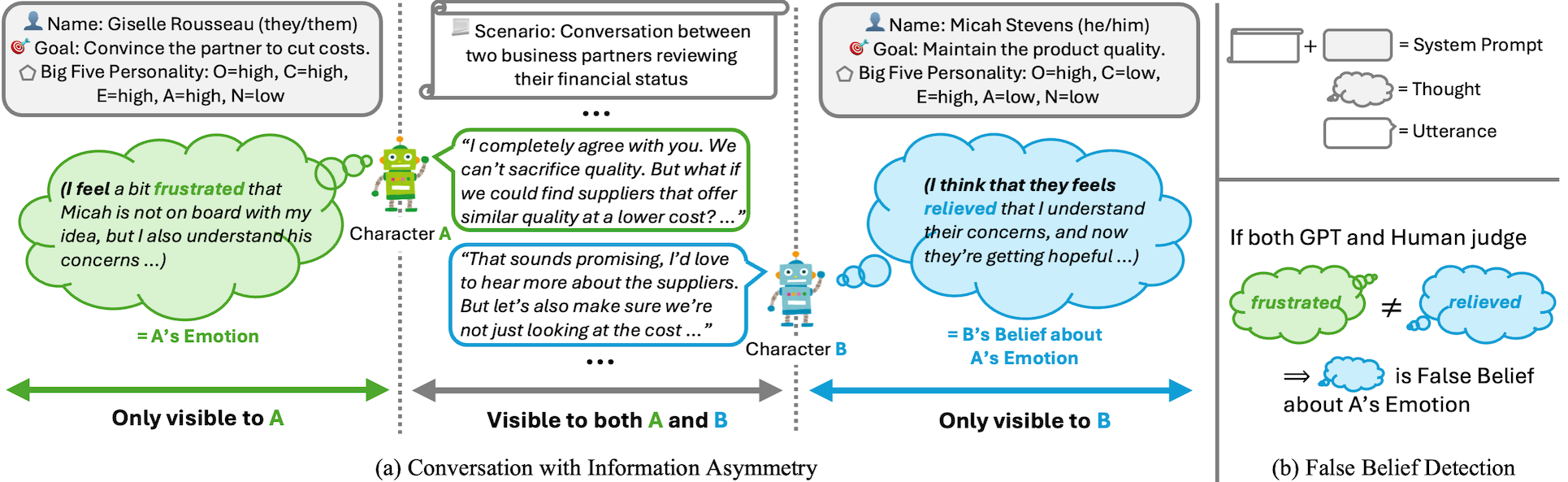}
    \caption{(a) Conversation between two role-playing LLMs with information asymmetry. Before speaking to the other, our Inner Speech prompting (e.g., I feel, or I think that he/she/they feels) promptes each agent to verbalize their first- and second-order mental states as thoughts. The verbalized thoughts are used as the answers to the questions in ToMATO. (b) To detect false beliefs, both GPT4o mini and human annotators judge whether character B misunderstands A's mental state at each turn.}
    \label{fig:is-overview}
\end{figure*}

To address the above issues, we introduce ToMATO, a new \textbf{T}heory-\textbf{o}f-\textbf{M}ind d\textbf{AT}aset generated via Inner Speech pr\textbf{O}mpting.\footnote{Our dataset and codes are available at \url{https://github.com/nttmdlab-nlp/ToMATO}.}
Firstly, ToMATO comprehensively evaluates first- and second-order ToM for five categories of mental states: beliefs, intentions, desires, emotions, and knowledge.
Secondly, we provide ToMATO-FB, a subset of ToMATO for evaluating understanding of false beliefs about the five mental states of others, e.g., understanding Bob thinks that Alice feels \textit{relieved}, while Alice feels \textit{frustrated} (Figure \ref{fig:is-overview}).
In this regard, ToMATO is the most comprehensive benchmark compared to the existing ones.
Lastly, ToMATO can evaluate the robustness of LLMs to the diverse personality traits of characters as seen in the real world.
See Table \ref{tab:tom-dataset} for the detailed comparison.

Collecting human conversations and mental states of participants with self-report can be challenging in terms of cost, privacy, and accuracy \cite{nisbett1977telling}.
As recent LLMs have been shown to role-play assigned personalities \cite{jiang-etal-2023-evaluating,jiang-etal-2024-personallm} and engage in conversations \cite{zhou2023sotopia}, we generate ToMATO with a newly designed LLM-LLM interaction.
Namely, we design Inner Speech prompting (Table \ref{tab:is-prompt}), which promotes role-playing LLMs to verbalize their mental states as thoughts in conversations with another LLM.
This idea is inspired by the debate in \citet{danica2023ai} about the feasibility of LLMs to simulate human participants in psychological science.
Moreover, we hypothesize that ensuring information asymmetry about thoughts is a crucial factor in having false beliefs about the mental states of others as shown in Figure \ref{fig:is-overview}.
In addition, we assign big five personality traits to LLMs to diversify utterances and thoughts.
This design enables ToMATO to evaluate robustness to diverse personality traits.
The effects of these approaches are verified with analyses in \S\ref{sec:analysis}.

We evaluate nine LLMs including local and proprietary ones on ToMATO.
Our experiments show that even the most advanced LLM, GPT-4o mini, under-performs the human performance in ToMATO.
In addition, we show that the ToM performance of LLMs drops for the false belief subset, ToMATO-FB.
Furthermore, we find that LLMs lack robustness to diverse personality traits.
These results suggest that ToM in LLMs is still far from deployable to real-world applications.

\begin{table*}[t]
    \centering
    \small
    \caption{Comparison of ToMATO to existing ToM benchmarks. ToMi \cite{le-etal-2019-revisiting}, Hi-ToM \cite{wu-etal-2023-hi}, BigToM \cite{gandhi2023understanding}, FauxPas-EAI \cite{shapira-etal-2023-well}, FANToM \cite{kim-etal-2023-fantom}, OpenToM \cite{xu2024opentom}, ToMBench \cite{chen2024tombench}. B: belief, I: intention, D: desire, E: emotion, K: knowledge, FB: false beliefs, W: world state, \#P: the number of personality trait patterns.}
    \begin{tabular}{c|c|c|c|c|c|c}
    \toprule
    & \multicolumn{3}{c|}{Assessable Mental State} &  & Input & Context \\
    Benchmark  & First-order & Second-order B about & FB about & \#P & Context &  Generator \\
    \midrule
    ToMi & B & B & W,B & - & Narrative & Template\\
    Hi-ToM & B & B & - & - & Narrative & Template\\
    BigToM & B & - & W & - & Narrative & Template+LLM\\
    FauxPas-EAI & B & - & - & - & Narrative & Psychological Test\\
    FANToM & B & B & B &  - & Conversation & Single LLM\\
    OpenToM & B,E & B & - & 3 & Narrative & Single LLM\\
    ToMBench & B,I,D,E,K & B & B & - & Narrative & Human\\
    \midrule
    ToMATO (ours) & B,I,D,E,K & B,I,D,E,K & B,I,D,E,K & 15 & Conversation & LLM-LLM Conversation\\
    \bottomrule
    \end{tabular}
    \label{tab:tom-dataset}
\end{table*}

\section{Preliminaries}

Theory of Mind (ToM) has been studied for decades \cite{premack_woodruff_1978}, with various definitions and conceptions proposed.
In this section, we clarify the scope addressed in this paper.

\paragraph{Mental states.} Following \citet{ma-etal-2023-towards-holistic}, we define the scope of ToM in this study by focusing on specific mental state categories identified by \citet{beaudoin2020systematic}: beliefs, intentions, desires, emotions, and knowledge.
Mental states are represented with mental (state) verbs \cite{shatz1983acquisition}, such as think, feel, and know.
Describing the remaining two categories defined by \citet{beaudoin2020systematic}, percepts and non-literal communications, require a multimodal context and/or pose challenges in verbalization.
Consequently, we have opted to exclude these two categories from our research scope.

\paragraph{First- and second-order mental states.}
First-order beliefs refer to one's beliefs about something, e.g., A thinks that X.
Second-order beliefs, on the other hand, often refer to one's beliefs about others' beliefs \cite{le-etal-2019-revisiting,sclar-etal-2023-minding}, e.g., B thinks that A thinks that Y.
We extend these notions to other mental states following previous studies on human ToM \cite{winner1991distinguishing,leekam1994autistic,hayashi2007young}.
Namely, we use the term first-order beliefs/intentions/desires/emotions/knowledge to refer to what A thinks/will/wants/feels/knows, and second-order beliefs about beliefs/intentions/desires/emotions/knowledge to refer to what B thinks that A thinks/will/wants/feels/knows.

\paragraph{False beliefs.} The false belief paradigm was initially introduced by \citet{wimmer1983beliefs}.
Understanding false beliefs of others, i.e., that others have wrong beliefs that differ from reality, has long been a prerequisite for ToM in humans \cite{wimmer1983beliefs} and machines \cite{le-etal-2019-revisiting}.
First-order false beliefs (FB) refer to beliefs about something that differs from reality, and second-order FB about beliefs refer to what B thinks that A thinks X, when A actually thinks Y.
In this study, we focus on second-order FB, which we simply call FB.
In human ToM, FB about a variety of mental states have been studied extensively \cite{gross1988false,shiverick2007second,smith2021preschoolers,wang2024picture}.
We also extend FB to the five mental states, i.e., FB about beliefs/intentions/desires/emotions/knowledge.
An example of FB about emotions is shown in Figure \ref{fig:is-overview}.

\section{Related Work}

\paragraph{Theory-of-Mind Benchmarks.}
LLMs have been reported to achieve human-level performance on various benchmarks \cite{openai2023gpt}.
In response to this trend, researchers have been gaining interest in whether LLMs have human-like ToM \cite{kosinski2023theory,bubeck2023sparks} or not \cite{ullman2023large,shapira-etal-2024-clever}.
To date, many benchmarks for evaluating ToM in machines have been constructed based on psychological tests designed for humans.
False belief tasks inspired by Sally-Anne test \cite{wimmer1983beliefs} have been widely used to evaluate understanding of wrong beliefs about object locations from narratives generated with templates \cite{nematzadeh-etal-2018-evaluating,le-etal-2019-revisiting,wu-etal-2023-hi,gandhi2023understanding}.
\citet{shapira-etal-2023-well} constructed a benchmark based on the faux pas test.
\citet{chen2024tombench} created ToMBench encompassing eight tasks known in psychological literature.
However, in addition to its potential to cause test set contamination due to the popularity of these tests \cite{shapira-etal-2024-clever}, these benchmarks are not aligned well with real-world scenarios in primarily the following aspects.

\paragraph{Assessable Mental States.} As discussed in \citet{ma-etal-2023-towards-holistic}, existing benchmarks evaluated only limited categories of mental states such as beliefs, even though humans infer other categories of mental states such as emotions or intentions of others in daily lives \cite{beaudoin2020systematic}.
In this regard, comprehensive ToM benchmarks are still lacking.
In particular, second-order ToM was primarily evaluated for beliefs as seen in Table \ref{tab:tom-dataset}, i.e., beliefs about mental states other than beliefs, such as emotions, have not been studied in machine ToM, even though they have been studied in human ToM \cite{gross1988false,smith2021preschoolers}.
Drawing conclusions about ToM in LLMs from such limited tests can induce media hype \cite{shapira-etal-2024-clever}.
In contrast, our benchmark, ToMATO, is aimed to comprehensively evaluate ToM reasoning about first- and second-order beliefs, intentions, desires, emotions, and knowledge.

\paragraph{False Beliefs.}
Existing ToM benchmarks often provide FB understanding tasks, but these are primarily focused on FB about beliefs of others or world states such as object locations \cite{le-etal-2019-revisiting,wu-etal-2023-hi,gandhi2023understanding,chen2024tombench}.
These FB are produced by leveraging information asymmetry between characters \cite{Braner2019Being}, which is caused by the physical movement of characters described in narratives.
\citet{kim-etal-2023-fantom} used a single LLM to generate multi-party conversations, where characters join or leave discussions to introduce information asymmetry about the topics.
In stark contrast, ToMATO-FB, the subset of ToMATO for evaluating FB about the five mental states, is constructed by our newly designed LLM-LLM conversations.
Notably, our LLM-LLM conversations involve information asymmetry about goals, personality traits, and thoughts of role-playing LLMs, whose effects are still under-studied in the context of LLM-LLM interactions \cite{zhou2024real}.
We argue that our design of LLM-LLM conversations is not only more aligned with real conversations, but also induces agents to frequently have FB about the mental states of others.
To the best of our knowledge, ToMATO is the first ToM benchmark generated via LLM-LLM conversations.
FB about the comprehensive mental states have not been explored in existing ToM benchmarks.

\paragraph{Personality Traits.} Even though personality traits are known to be correlated with mental states \cite{costa1980influence,izard1993stability,lucas2001understanding,kashdan2010psychological} and language use \cite{norman1963toward,mehl2006personality} in psychological studies, in most ToM benchmarks, correct predictions can be made without considering personality traits of characters.
While OpenToM \cite{xu2024opentom} introduced characters with three personality traits (considerate, inconsiderate, and negativistic), ToMATO covers 15 patterns of big five personality traits.
See Figure \ref{fig:is-overview} for examples.

\begin{table}[tbp]
    \centering
    \small
    \caption{Inner Speech prompts for each type of mental states.}
    \begin{tabular}{c|ll}
    \toprule
    Mental State & \multicolumn{2}{c}{Inner Speech Prompt}\\
     $T$ & \multicolumn{1}{c}{$p_{\rm IS}^{T_1}$} & \multicolumn{1}{c}{$p_{\rm IS}^{T_2}$} \\
    \midrule
    Belief & (I think & (I think that he/she thinks\\
    Intention & (I will & (I think that he/she will \\
    Desire & (I want & (I think that he/she wants \\
    Emotion & (I feel & (I think that he/she feels \\
    Knowledge & (I know & (I think that he/she knows\\
    \bottomrule
    \end{tabular}
    \label{tab:is-prompt}
\end{table}

\paragraph{Input Context.}
In addition, most benchmarks evaluate ToM with narratives as input as shown in Table \ref{tab:tom-dataset}.
FANToM \citet{kim-etal-2023-fantom} adopts conversations generated by a single LLM as input for the first time to reduce reporting bias \cite{gordon2013reporting} and align with real-world scenarios.
While our ToMATO also employs conversations as input, the conversations and thoughts are generated by role-playing LLMs with distinct personality traits assigned.

\section{ToMATO Benchmark}
\label{sec:tomato}
In this section, we describe the overview of our ToMATO benchmark: automatic construction process with LLMs, quality validation, and its statistics.
Following the success of existing studies \cite{kim-etal-2023-soda,kim-etal-2023-fantom}, we also use LLMs to generate conversations.
We employ Llama-3-70B-Instruct \cite{dubey2024llama} because of its transparency and relatively high scores on popular benchmarks \cite{chiang2024chatbot}.\footnote{Proprietary LLMs such as OpenAI's ChatGPT are not transparent. Actually, ChatGPT may use internal system prompts \cite{tweet2024system}. Moreover, they are continuously updated and previous LLMs would be unavailable, which impairs reproducibility.}

\subsection{Notation}
The ToMATO benchmark is formulated as a multiple-choice question answering task due to its reliable evaluation.
Each instance in the benchmark includes conversation $C$, question $Q$, four options $O=\{o_i\}_{i=1}^4$ as input and ground-truth answer $A \in O$.
Let $\pi_A$ and $\pi_B$ be role-playing LLMs with the multi-turn conversation capability that serve as characters $A$ and $B$ in conversation, respectively.
Conversation $C_{1:N}$ consists of utterances $\{u_1^A, u_1^B, ..., u_N^A, u_N^B\}$, where $u_i^A$ is the $i$-th utterance of character $A$.
We define the category of mental states as $T$.
The actual first- and second-order mental state of character $A$ for type $T$ when $A$ says $u_i^A$ is defined as $m_i^{A,T_1}$ and $m_i^{A,T_2}$, respectively.
$p_{\rm SY}$ and $p_{\rm IS}$ represent the system prompt and the proposed inner speech prompt, respectively, which are explained in the following sections.

\subsection{System Prompt}
We design system prompts to guide LLM-LLM conversations, extending SOTOPIA \cite{zhou2023sotopia} to consider the big five personality traits.
SOTOPIA was initially proposed to evaluate social interaction of LLMs in LLM-LLM interactions, providing conversation scenarios from eight categories, such as persuasion, and character profiles.
We sample 160 conversation scenarios uniformly from the eight categories, two characters for each conversation, and their goals, from SOTOPIA.
See Figure \ref{fig:is-overview} for examples.
Then, we sample five pairs of characters for each scenario to prepare control conditions with regard to personality traits, resulting in 800 conversations.
This design enables ToMATO to evaluate the robustness to diverse personality traits.
In detail, we extend the naive prompt \cite{jiang-etal-2023-evaluating}, which reflects only one factor (e.g., You are \{an open/a closed\} person.), to include a combination of five factors of big five personality traits \cite{de2000big} (e.g., You are \{an open / a closed\}, \{conscientious / unconscientious\}, \{extraversive / introversive\}, \{agreeable / disagreeable\}, and \{neurotic / stable\} person.).
We compile the above information to formulate system prompts, $p_{\rm SY}^A$, given to $\pi_A$.

\subsection{Inner Speech Prompting}

In order to make the inherently unobservable mental states observable, we propose Inner Speech (IS) prompting.
IS prompting promotes role-playing LLMs to verbalize their subjective mental states in conversations.
Since IS prompting can verbalize mental states as thoughts at any point during the conversation, ToMATO can also evaluate the understanding of dynamic changes in mental states.
The actual IS prompts are given in Table \ref{tab:is-prompt}.
Moreover, IS prompting can generate first- and second-order mental states for five types by adjusting prompts.
In order to ensure that the output follows the format, (\{thought\}) ``\{utterance\}'', IS prompting specifies the prefix of the output, and LLMs generate the continuation.
This format design enables deleting only thoughts with regular expressions in the next section.

\begin{table*}[htbp]
\footnotesize
\centering
\caption{ToM Performance on ToMATO (\%). B: belief, I: intention, D: desire, E: emotion, K: knowledge.
LO: lexical overlap baseline. FB: false belief tasks. The mean$\pm$standard deviations over five runs are reported.}
\begin{tabular}{cc|c|ccccccccc|c}
\toprule
 \multicolumn{2}{c|}{Mental} &  & \multicolumn{2}{c}{Llama3} & \multicolumn{2}{c}{Llama3.1} & Gemma2 & Mistral & Mixtral & \multicolumn{2}{c|}{GPT} & \\
 \multicolumn{2}{c|}{State} & LO & 8B & 70B & 8B & 70B & 9B & 7B & 8x7B & 3.5-Trubo & 4o mini & Human \\\midrule
  & 1st & 40.8 & 53.1{\scriptsize$\pm$1.4} & 81.5{\scriptsize$\pm$0.4} & 61.9{\scriptsize$\pm$1.8} & 82.1{\scriptsize$\pm$0.4} & 79.2{\scriptsize$\pm$0.2} & 62.0{\scriptsize$\pm$0.4} & 64.0{\scriptsize$\pm$0.3} & 58.7{\scriptsize$\pm$0.6} & 76.3{\scriptsize$\pm$0.7} & 87.5\\
B & 2nd & 38.0 & 37.6{\scriptsize$\pm$0.7} & 68.1{\scriptsize$\pm$0.8} & 40.9{\scriptsize$\pm$0.7} & 69.7{\scriptsize$\pm$0.5} & 68.5{\scriptsize$\pm$0.3} & 51.0{\scriptsize$\pm$0.6} & 52.4{\scriptsize$\pm$0.3} & 49.9{\scriptsize$\pm$1.2} & 65.2{\scriptsize$\pm$0.7} & 87.5\\
 & FB & 37.1 & 34.7{\scriptsize$\pm$0.8} & 60.1{\scriptsize$\pm$1.4} & 38.7{\scriptsize$\pm$0.6} & 62.8{\scriptsize$\pm$1.2} & 61.2{\scriptsize$\pm$0.6} & 42.5{\scriptsize$\pm$1.3} & 43.5{\scriptsize$\pm$0.7} & 43.1{\scriptsize$\pm$1.7} & 60.2{\scriptsize$\pm$1.4} & 84.4\\\midrule
 & 1st & 35.0 & 56.4{\scriptsize$\pm$0.7} & 85.0{\scriptsize$\pm$0.5} & 67.0{\scriptsize$\pm$1.2} & 85.6{\scriptsize$\pm$0.6} & 80.6{\scriptsize$\pm$0.4} & 67.9{\scriptsize$\pm$0.4} & 64.8{\scriptsize$\pm$0.3} & 56.6{\scriptsize$\pm$0.6} & 80.1{\scriptsize$\pm$0.3} & 96.9\\
I & 2nd & 35.5 & 41.9{\scriptsize$\pm$0.7} & 71.2{\scriptsize$\pm$0.5} & 47.3{\scriptsize$\pm$1.5} & 69.6{\scriptsize$\pm$0.5} & 65.8{\scriptsize$\pm$0.6} & 56.8{\scriptsize$\pm$0.6} & 58.6{\scriptsize$\pm$0.2} & 49.4{\scriptsize$\pm$0.8} & 64.9{\scriptsize$\pm$0.5} & 93.8\\
 & FB & 32.8 & 29.8{\scriptsize$\pm$2.7} & 57.4{\scriptsize$\pm$0.7} & 36.9{\scriptsize$\pm$0.9} & 53.6{\scriptsize$\pm$1.7} & 48.2{\scriptsize$\pm$1.4} & 40.7{\scriptsize$\pm$1.1} & 42.3{\scriptsize$\pm$1.2} & 35.2{\scriptsize$\pm$1.7} & 47.4{\scriptsize$\pm$0.8} & 78.1\\\midrule
 & 1st & 32.0 & 60.1{\scriptsize$\pm$1.3} & 86.1{\scriptsize$\pm$0.3} & 74.0{\scriptsize$\pm$1.4} & 88.7{\scriptsize$\pm$0.7} & 86.3{\scriptsize$\pm$0.3} & 74.7{\scriptsize$\pm$0.3} & 73.9{\scriptsize$\pm$0.4} & 69.8{\scriptsize$\pm$0.3} & 81.9{\scriptsize$\pm$0.6} & 93.8\\
D & 2nd & 37.9 & 43.4{\scriptsize$\pm$0.9} & 75.6{\scriptsize$\pm$0.5} & 50.7{\scriptsize$\pm$1.7} & 79.5{\scriptsize$\pm$0.8} & 75.2{\scriptsize$\pm$0.3} & 58.2{\scriptsize$\pm$0.4} & 60.6{\scriptsize$\pm$0.1} & 55.4{\scriptsize$\pm$0.9} & 75.7{\scriptsize$\pm$0.4} & 84.4\\
 & FB & 39.2 & 34.9{\scriptsize$\pm$1.5} & 67.2{\scriptsize$\pm$1.0} & 43.0{\scriptsize$\pm$4.2} & 75.9{\scriptsize$\pm$2.4} & 72.2{\scriptsize$\pm$0.4} & 48.9{\scriptsize$\pm$0.8} & 47.3{\scriptsize$\pm$0.5} & 47.1{\scriptsize$\pm$0.6} & 71.8{\scriptsize$\pm$0.9} & 78.1\\\midrule
 & 1st & 35.6 & 56.9{\scriptsize$\pm$1.3} & 80.4{\scriptsize$\pm$0.3} & 64.1{\scriptsize$\pm$1.3} & 82.2{\scriptsize$\pm$0.5} & 79.0{\scriptsize$\pm$0.6} & 60.8{\scriptsize$\pm$0.4} & 60.2{\scriptsize$\pm$0.3} & 61.3{\scriptsize$\pm$1.1} & 77.2{\scriptsize$\pm$0.6} & 93.8\\
E & 2nd & 28.5 & 44.5{\scriptsize$\pm$0.7} & 74.0{\scriptsize$\pm$0.4} & 51.3{\scriptsize$\pm$1.1} & 74.8{\scriptsize$\pm$0.4} & 76.6{\scriptsize$\pm$0.6} & 57.8{\scriptsize$\pm$0.7} & 58.5{\scriptsize$\pm$0.5} & 50.7{\scriptsize$\pm$0.5} & 71.9{\scriptsize$\pm$0.8} & 81.2\\
 & FB & 29.1 & 36.5{\scriptsize$\pm$2.0} & 71.0{\scriptsize$\pm$1.5} & 47.2{\scriptsize$\pm$5.0} & 69.1{\scriptsize$\pm$0.9} & 71.7{\scriptsize$\pm$1.0} & 48.0{\scriptsize$\pm$0.9} & 54.6{\scriptsize$\pm$0.6} & 37.5{\scriptsize$\pm$0.8} & 72.0{\scriptsize$\pm$0.9} & 71.9\\\midrule
 & 1st & 42.3 & 47.2{\scriptsize$\pm$1.0} & 73.5{\scriptsize$\pm$0.5} & 53.7{\scriptsize$\pm$2.0} & 74.8{\scriptsize$\pm$0.8} & 74.7{\scriptsize$\pm$0.4} & 62.5{\scriptsize$\pm$0.4} & 63.5{\scriptsize$\pm$0.5} & 56.1{\scriptsize$\pm$0.9} & 73.3{\scriptsize$\pm$0.2} & 96.9\\
K & 2nd & 40.2 & 36.9{\scriptsize$\pm$0.6} & 66.6{\scriptsize$\pm$0.7} & 44.6{\scriptsize$\pm$1.2} & 73.6{\scriptsize$\pm$0.7} & 70.3{\scriptsize$\pm$0.2} & 59.1{\scriptsize$\pm$0.6} & 58.9{\scriptsize$\pm$0.3} & 53.0{\scriptsize$\pm$0.6} & 69.6{\scriptsize$\pm$0.4} & 87.5\\
 & FB & 46.3 & 27.8{\scriptsize$\pm$1.7} & 58.0{\scriptsize$\pm$1.0} & 36.8{\scriptsize$\pm$3.8} & 63.6{\scriptsize$\pm$1.8} & 59.3{\scriptsize$\pm$0.4} & 51.1{\scriptsize$\pm$1.1} & 46.3{\scriptsize$\pm$0.6} & 45.7{\scriptsize$\pm$1.0} & 58.6{\scriptsize$\pm$0.6} & 93.8\\\midrule
 \multicolumn{2}{c|}{ALL} & 36.8 & 47.5{\scriptsize$\pm$0.2} & 76.0{\scriptsize$\pm$0.2} & 55.2{\scriptsize$\pm$0.3} & 77.9{\scriptsize$\pm$0.2} & 75.4{\scriptsize$\pm$0.1} & 60.9{\scriptsize$\pm$0.1} & 61.4{\scriptsize$\pm$0.1} & 55.8{\scriptsize$\pm$0.3} & 73.5{\scriptsize$\pm$0.1} & 87.3\\
\bottomrule
\end{tabular}
\label{tab:main-results}
\end{table*}

\subsection{Conversation with Information Asymmetry}
At each turn, each agent is prompted to generate utterance $u_i$ and its mental state $m_i^T$ as follows:
\begin{align*}
    u_i^A, & m_i^{A,T_1} \sim \pi_A(u, m|p_{\rm SY}^A, C_{1:i-1}, p_{\rm IS}^{T_1}),\\
    u_i^B, & m_i^{B,T_2} \sim \pi_B(u, m|p_{\rm SY}^B, C_{1:i-1}, u_i^A, p_{\rm IS}^{T_2}),\\
    & {\rm where} ~~ C_{1:i-1} = \{u_1^A, u_1^B, ..., u_{i-1}^A, u_{i-1}^B\}.
\end{align*}
This sampling process continues until the $N$-th turn.
Then, we obtain $2N$ utterances with corresponding first- and second-order mental states for type $T$.
We repeat this multi-turn conversation for each mental state, $T\in\{\textrm{Belief, Intention, Desire, Emotion, Knowledge}\}$, each scenario, and each pair of characters.
$N$ is set seven because we found longer conversations tended to be redundant.

Here, by hiding one's mental states from the other, we ensure information asymmetry about thoughts between the two as in human conversations.
To delete thoughts from the outputs, we instruct LLMs to include their thoughts in ``()'', which can be detected with regular expressions, in the system prompts.
We also make the goal and personality of one agent in the system prompt invisible to the other.
This information asymmetry has a positive effect on generating false beliefs as shown in \S\ref{sec:analysis}.

\subsection{Multiple-choice QA Dataset Construction}
The generated conversation and mental states are converted into a multiple-choice QA dataset, consisting of conversation $C$, question $Q$, options $O$, and ground-truth answer $A$.
Conversation $C$ is generated in the former multi-turn conversation.
Question $Q$ asks about the mental state $T$ of a character at $i$-th turn in $C$.
$Q$ is generated with predefined templates for each utterance $u_i$ in $C$, and $A$ is the thought $m_i$ corresponding to the utterance.

For collecting incorrect options in $O$, we randomly sampled three options from $\{m_i^A\}_{i=1}^N\setminus\{A\}$ for each question.
We do so because intentionally creating incorrect options can introduce spurious correlations, e.g., manually written incorrect options tend to be shorter than correct ones \cite{guo-etal-2023-desiq}.
The effect of the incorrect option sampling on word-level spurious correlations is discussed in \S\ref{sec:analysis}.

\subsection{False Belief Detection}

We build ToMATO-FB, the second-order false belief subset of ToMATO, by comparing first-order mental state of A, $m_i^{A,T_1}$, and second-order mental state of B, $m_i^{B,T_2}$, at each turn with human and LLM judges.
Namely, we instruct three human annotators and GPT-4o mini to judge whether B correctly infers A's mental state or not.
When both the majority of annotators and GPT-4o mini agree that B partially misunderstands A's mental state, it is added to ToMATO-FB.

\subsection{Quality Validation \& Statistics}
\paragraph{Validation.}
We validate the quality of ToMATO using Amazon Mechanical Turk (MTurk).
First, the consistency and harmlessness of conversations are verified by three qualified annotators, following \citet{kim-etal-2023-fantom}.
This is to judge whether the generated conversations are suitable as input.
Conversations flagged by the majority (5.8\%) are excluded from the benchmark.
Then, we verify whether the correct and incorrect options are indeed correct and incorrect for each question following \citet{zadeh-etal-2019-socialiq,kim-etal-2023-fantom}.
We use both MTurk and GPT-4o mini to strictly verify the quality of the questions in ToMATO.
Then, those deemed valid by both the majority of annotators and GPT-4o mini are included in ToMATO.

\begin{table}[tbp]
    \centering
    \small
    \caption{\#C: the num. of conversations. \#Q: the num. of questions. Avg. \#Token: the average num. of tokens in $u$. Avg. \#Utt: the average num. of utterances in $C$.}
    \begin{tabular}{ccccc}
    \toprule
    & \#C & \#Q & Avg. \#Token & Avg. \#Utt \\
    \midrule
    FANToM & 256 & 10k & 31.4 & 13.8 \\
    ToMATO & 753 & 5.4k & 41.6 & 16 \\
    \bottomrule
    \end{tabular}
    \label{tab:statistics-comparison}
\end{table}

\paragraph{Statistics.}
After removing invalid instances, the ToMATO benchmark contains 5.4k questions and 753 conversations.
ToMATO-FB consists of 806 questions.
Table \ref{tab:statistics-comparison} compares statistics of ToMATO and FANToM \cite{kim-etal-2023-fantom}.

\begin{table*}[htbp]
\footnotesize
\centering
\caption{First-order ToM Performance for each factor of big five personality traits of characters. For each factor of big five (O=openness to experience, C=conscientiousness, E=extraversion, A=agreeableness, N=neuroticism), the scores$\pm$standard deviations on two subsets (the corresponding factor is high and low) averaged over five runs are reported.}
\begin{tabular}{cc|c|ccccccccc}
\toprule
\multicolumn{2}{c|}{Big} &  & \multicolumn{2}{c}{Llama-3} & \multicolumn{2}{c}{Llama-3.1} & \multicolumn{1}{c}{Gemma-2} & \multicolumn{1}{c}{Mistral} & \multicolumn{1}{c}{Mixtral} & \multicolumn{2}{c}{GPT} \\
\multicolumn{2}{c|}{Five} & LO & 8B & 70B & 8B & 70B & 9B & 7B & 8x7B & 3.5-Trubo & 4o mini \\\midrule
\multirow{2}{*}{O} & high  & 37.3 & 54.8{\scriptsize$\pm$0.4} & 81.2{\scriptsize$\pm$0.3} & 64.1{\scriptsize$\pm$1.3} & 82.4{\scriptsize$\pm$0.4} & 80.1{\scriptsize$\pm$0.2} & 65.2{\scriptsize$\pm$0.2} & 64.3{\scriptsize$\pm$0.1} & 60.5{\scriptsize$\pm$0.2} & 77.2{\scriptsize$\pm$0.2}\\
& low  & 37.4 & 54.0{\scriptsize$\pm$0.3} & 81.1{\scriptsize$\pm$0.2} & 63.2{\scriptsize$\pm$0.8} & 82.6{\scriptsize$\pm$0.4} & 79.2{\scriptsize$\pm$0.3} & 65.6{\scriptsize$\pm$0.4} & 66.0{\scriptsize$\pm$0.1} & 59.4{\scriptsize$\pm$0.6} & 78.2{\scriptsize$\pm$0.2}\\\midrule
\multirow{2}{*}{C} & high  & 37.9 & 56.6{\scriptsize$\pm$0.4} & 82.4{\scriptsize$\pm$0.2} & 65.5{\scriptsize$\pm$0.6} & 83.3{\scriptsize$\pm$0.2} & 80.0{\scriptsize$\pm$0.2} & 66.3{\scriptsize$\pm$0.2} & 66.8{\scriptsize$\pm$0.1} & 61.5{\scriptsize$\pm$0.4} & 78.7{\scriptsize$\pm$0.2}\\
& low  & 36.3 & 50.3{\scriptsize$\pm$0.3} & 78.7{\scriptsize$\pm$0.2} & 60.1{\scriptsize$\pm$1.5} & 80.8{\scriptsize$\pm$0.3} & 79.0{\scriptsize$\pm$0.4} & 63.5{\scriptsize$\pm$0.6} & 61.5{\scriptsize$\pm$0.1} & 57.1{\scriptsize$\pm$0.2} & 75.5{\scriptsize$\pm$0.5}\\\midrule
\multirow{2}{*}{E} & high  & 37.7 & 54.1{\scriptsize$\pm$0.7} & 82.4{\scriptsize$\pm$0.2} & 64.7{\scriptsize$\pm$0.8} & 84.1{\scriptsize$\pm$0.4} & 81.4{\scriptsize$\pm$0.2} & 66.2{\scriptsize$\pm$0.3} & 65.7{\scriptsize$\pm$0.2} & 60.3{\scriptsize$\pm$0.6} & 78.8{\scriptsize$\pm$0.4}\\
& low  & 37.1 & 54.8{\scriptsize$\pm$0.4} & 79.9{\scriptsize$\pm$0.3} & 62.8{\scriptsize$\pm$1.0} & 81.0{\scriptsize$\pm$0.2} & 78.2{\scriptsize$\pm$0.2} & 64.6{\scriptsize$\pm$0.2} & 64.4{\scriptsize$\pm$0.2} & 59.7{\scriptsize$\pm$0.4} & 76.5{\scriptsize$\pm$0.2}\\\midrule
\multirow{2}{*}{A} & high  & 38.8 & 55.0{\scriptsize$\pm$0.8} & 83.4{\scriptsize$\pm$0.3} & 65.7{\scriptsize$\pm$0.6} & 85.0{\scriptsize$\pm$0.3} & 82.5{\scriptsize$\pm$0.2} & 67.2{\scriptsize$\pm$0.3} & 65.7{\scriptsize$\pm$0.2} & 60.7{\scriptsize$\pm$0.7} & 79.3{\scriptsize$\pm$0.4}\\
& low  & 36.2 & 54.0{\scriptsize$\pm$0.4} & 79.3{\scriptsize$\pm$0.2} & 62.2{\scriptsize$\pm$1.1} & 80.4{\scriptsize$\pm$0.5} & 77.5{\scriptsize$\pm$0.2} & 63.9{\scriptsize$\pm$0.1} & 64.5{\scriptsize$\pm$0.2} & 59.5{\scriptsize$\pm$0.4} & 76.3{\scriptsize$\pm$0.2}\\\midrule
\multirow{2}{*}{N} & high  & 34.7 & 47.7{\scriptsize$\pm$1.2} & 78.8{\scriptsize$\pm$0.2} & 59.4{\scriptsize$\pm$1.6} & 81.3{\scriptsize$\pm$0.8} & 77.4{\scriptsize$\pm$0.4} & 62.0{\scriptsize$\pm$0.8} & 59.2{\scriptsize$\pm$0.3} & 56.2{\scriptsize$\pm$0.6} & 75.5{\scriptsize$\pm$0.8}\\
& low  & 37.9 & 55.9{\scriptsize$\pm$0.2} & 81.6{\scriptsize$\pm$0.3} & 64.6{\scriptsize$\pm$0.8} & 82.7{\scriptsize$\pm$0.4} & 80.2{\scriptsize$\pm$0.2} & 66.1{\scriptsize$\pm$0.1} & 66.3{\scriptsize$\pm$0.1} & 60.8{\scriptsize$\pm$0.4} & 78.1{\scriptsize$\pm$0.1}\\
\bottomrule
\end{tabular}
\label{tab:big-five-first}
\end{table*}

\section{Experiments}
We evaluate ToM in LLMs on ToMATO, exploring whether our approach uncovers insights into ToM in current LLMs with regard to various mental states, false beliefs, and personality traits that were not attainable with previous datasets.

\subsection{Experimental Setup}
\paragraph{Baselines.}
We evaluated nine LLMs: Llama-3-Instruct (8B and 70B) , Llama-3.1-Instruct(8B and 70B) \cite{dubey2024llama}, Gemma-2-IT (9B) \cite{gemma_2024}, Mistral-Instruct (7B), Mixtral-8x7B-Instruct, GPT-3.5-Turbo, and GPT-4o mini \cite{gpt-4o-mini}.
For the local LLMs, 4-bit quantization with bitsandbytes\footnote{\url{https://github.com/bitsandbytes-foundation/bitsandbytes}} was used for inference.
We employed lexical overlap (LO) as a naive baseline.
LO simply selects the options that have the most words in common with the questions \cite{shinoda-etal-2023-which}.\footnote{Note that the random baseline is 25\% in ToMATO.}

\paragraph{Human Baseline.}
We also measured the human performance using MTurk.
Annotators who are awarded Masters Qualification solved 32 questions for each subset, i.e., 480 questions in total.

\subsection{Experimental Results}
\paragraph{Do LLMs have human-level ToM?}
Table \ref{tab:main-results} shows the results of LLMs and the human baseline.
The results showed that even the most advanced LLMs, such as GPT-4o mini and Llama-3.1 70B, lag behind the human baseline.
We also tested Chain-of-Thought prompting and fine-tuning, but they were not sufficient to achieve human-level performance.

Among the baseline LLMs, Llama-3.1-70B-Instruct was the state-of-the-art.
However, because the ToMATO benchmark was generated with Llama-3-70B-Instruct, it is unfair to compare Llama models with other LLMs.
This is one of the limitations in constructing benchmarks with LLMs.
Among the small language models, surprisingly, Gemma-2-9B-it achieved the highest scores, which is comparable to GPT-4o mini.
Knowledge distillation from larger language models, used to train Gemma-2-9B \cite{gemma_2024} but not Llama-3-8B \cite{dubey2024llama}, might be the key to the high performance despite its small size.

\begin{table}[tbp]
    \centering
    \small
    \caption{Ablation study of information asymmetry on the frequency probability (\%) of false beliefs.}
    \begin{tabular}{c|c|c|c}
    \toprule
    \multicolumn{2}{c|}{Information Asymmetry} & \multicolumn{2}{c}{Judge} \\
    System Prompt & Thought & GPT & Human \\
    \midrule
    \checkmark & \checkmark & \textbf{46.6} & \textbf{51.0} \\
     & \checkmark & 40.4 & 32.0 \\
    \checkmark &  & 46.0 & 32.0 \\
     &  & 39.0 & 30.5\\
     \bottomrule
    \end{tabular}
    \label{tab:ablation-info-asymmetry}
\end{table}

\paragraph{Does the ToM performance vary depending on the mental state?} For LLMs, desires are relatively easy to understand, and knowledge is hard to infer among the five mental states in ToMATO.
Interestingly, understanding desires is easier than beliefs for LLMs, which is consistent with children \cite{repacholi1997early,rakoczy2007way}.
We also showed that second-order mental states, especially for the ToMATO-FB subset, are consistently challenging for every mental state category.
This is also the case in human ToM for beliefs \cite{perner1985john}.
These insights were found for the first time due to the comprehensiveness of ToMATO. 
Thus, ToMATO would be useful to precisely understand the limitations of ToM in LLMs and gain insights into the directions toward human-like ToM.
For example, as done with children \cite{hughes1998understanding}, it is feasible to track the development of ToM in LLMs for each mental state with ToMATO during training.

\begin{figure}[tbp]
    \centering
    \begin{tabular}{cc}
    \includegraphics[width=3.7cm]{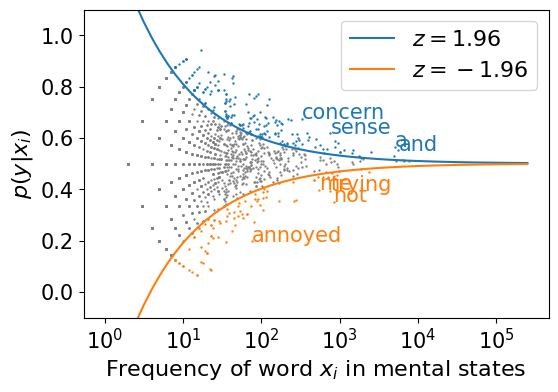} & \includegraphics[width=3.7cm]{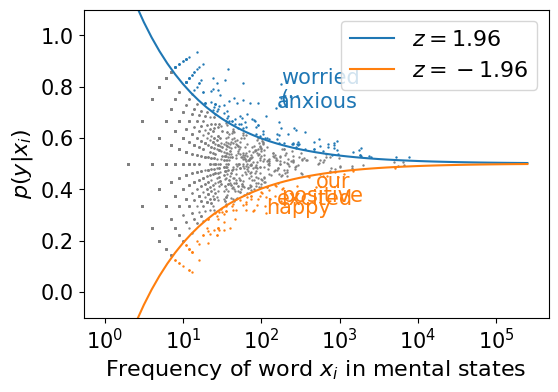} \\
    (a) Agreeable & (b) Neurotic \\
    \end{tabular}
    \caption{Statistical word-level correlation analysis \cite{gardner-etal-2021-competency} between the generated thoughts and the personality traits given in system prompts.}
    \label{fig:z-stat-personality}
\end{figure}

\def\WidthZstat{4cm}

\begin{figure*}[htbp]
    \centering
    \begin{tabular}{cccc}
    \includegraphics[width=\WidthZstat]{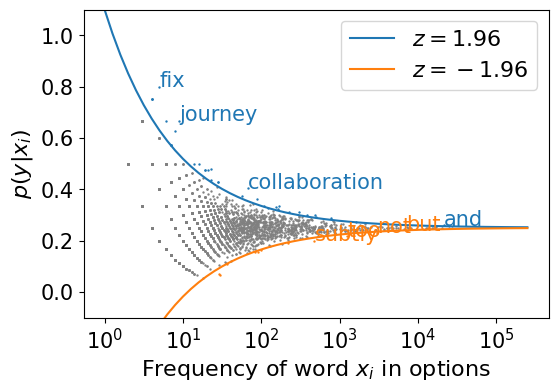} & \includegraphics[width=\WidthZstat]{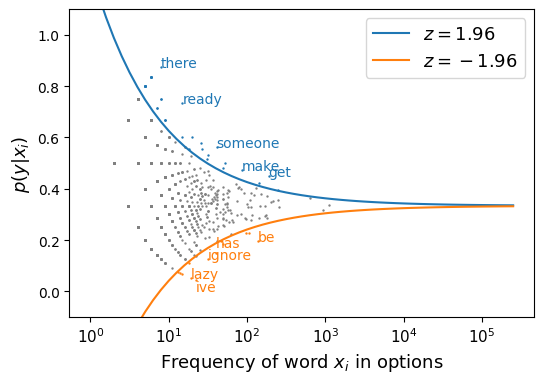} & \includegraphics[width=\WidthZstat]{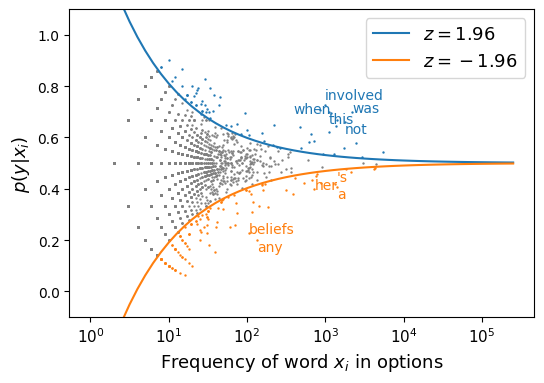} & \includegraphics[width=\WidthZstat]{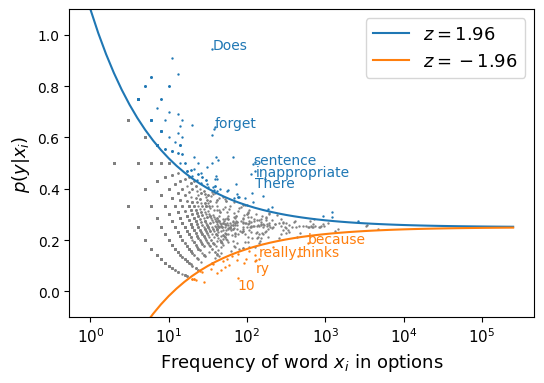} \\
    (a) ToMATO (ours) & (b) Social-IQa & (c) FANToM & (d) ToMBench \\
    \end{tabular}
    \caption{Statistical word-level correlation analysis \cite{gardner-etal-2021-competency} on four benchmarks. Among the four, ToMATO (ours) contains the fewest word-level spurious correlations in options, indicating sophisticated solutions are needed to achieve higher scores than the random baseline on ToMATO.}
    \label{fig:z-statistics}
\end{figure*}

\paragraph{Is ToM in LLMs robust to diverse personality traits?}
Table \ref{tab:big-five-first} shows the first-order ToM performance for each factor of big five.
E.g., for openness to experience (O), we split ToMATO into questions asking about characters with open (O=high) and closed (O=low) personalities, and reported the average scores for the two subsets.
The results showed that the performance varied based on the personality traits of the characters.
Namely, LLMs tended to degrade the performance of understanding mental states of unconscientious (C=low), introversive (E=low), disagreeable (A=low), or neurotic (N=high) characters.
The scores for E=high are higher than for E=low possibly because extraversive persons tend to express their emotions \cite{riggio2002emotional}.
We argue that the robustness of ToM to various personality traits should be improved for deploying ToM in LLMs to real-world applications, as humans possess diverse personalities.

\paragraph{Do LLMs exploit shortcut solutions?}
Most LLMs achieved higher scores for first- and second-order mental states than the LO baseline.
This indicates that the LLMs do not rely solely on shortcut solutions based on LO.
However, for ToMATO-FB, smaller LLMs performed worse than LO in some cases.
This indicates that understanding false beliefs remains a fundamental challenge for current LLMs.

\section{Analysis on the ToMATO benchmark}
\label{sec:analysis}

\paragraph{Is information asymmetry about thoughts effective for generating false beliefs?}
We conjecture that information asymmetry about their thoughts, goals, and personality traits between two LLMs in conversations is a key factor in inducing false beliefs about the mental states of the other.
To verify this hypothesis, we conducted ablation studies for the generation process.
Namely, we investigated the effect of the invisibility of one's thoughts and system prompts including goals and personality traits to the other on the frequency probability of false belief generation.
We evaluated 3k instances with GPT-4o mini and 200 instances with three human annotators of MTurk for each generation process.
We used majority vote to aggregate the human annotations.
Results are given in Table \ref{tab:ablation-info-asymmetry}.
The results showed that information asymmetry about both system prompts and thoughts encourages false belief generation.

\paragraph{Does ToMATO reflect personality traits given in prompts?}
To answer this question, we conducted the z-statistics analysis \cite{gardner-etal-2021-competency} for the correlations between the output tokens and the personality traits given in the prompts.
We first sampled one scenario from each category in SOTOPIA and generated conversations and thoughts with our approach in \S\ref{sec:tomato}.
We assigned every pattern of the big five personality, i.e., $32=2^5$ patterns in total, to one agent for each scenario.

Some results are displayed in Figure \ref{fig:z-stat-personality}.
Here, $y$ denotes the probability of word $x_i$ to appear in the output when the corresponding personality specified in prompts is high.
The colored tokens above or below the curves are significantly positively or negatively correlated to the assigned personality factor.
This indicates that the big five personality factors given in prompts have intentionally affected the generation.
E.g., agents who are assigned neurotic often generate ``worried'' and those who are assigned not neurotic often generate ``happy'' in their thoughts.

We also conducted a pairwise comparison, following \citet{jiang-etal-2023-evaluating}, to see if the specified personality traits are reflected properly with MTurk and GPT-4o mini.
We showed that 70-80\% of the outputs reflect the specified personality traits for O, E, A, and N.
Among the five, C is less reflected as intended, which is consistent with \citet{jiang-etal-2023-evaluating}.
Inducing conscientiousness in outputs is future work.

\paragraph{Can ToMATO be easily solved with shortcut solutions?}
Language understanding benchmarks should not be easily solved with shortcut solutions based on spurious correlations to ensure that those benchmarks measure intended abilities \cite{sugawara-tsugita-2023-degrees}.
In general, multiple-choice QA datasets often suffer from spurious correlations such as word-label correlation, and lexical overlap \cite{Yu2020ReClor,shinoda-etal-2023-which}.
First, for lexical overlap, the LO baseline does not achieve high performance compared to the human baseline as shown in Table \ref{tab:main-results}.

Second, for word-label correlation, we again conducted the z-statistics analysis \cite{gardner-etal-2021-competency} to identify statistically significant correlations between words in options and binary labels, for four benchmarks including ToMATO.
In z-statistics analysis, the frequencies and probabilities of each word that appears in correct options are plotted as shown in Figure \ref{fig:z-statistics}.
When the probabilities of words that appear in correct options are significantly higher ($z \geq 1.96$) or lower ($z \leq - 1.96$) than the random baseline, the words are colored.
In detail, the ratios (\%) of the number of biased (colored) words in options to the vocabulary size are 1.16, 3.34, 4.49, and 6.04 for ToMATO, Social-IQa, FANToM, and ToMBench, respectively.
These results indicate that ToMATO contains the fewest word-level spurious correlations among the four benchmarks.
Based on these analyses, we claim that ToMATO is so challenging that it requires models to acquire more sophisticated solutions than shortcuts to achieve human-level performance.

\section{Conclusion}
A comprehensive evaluation of ToM using our ToMATO would be valuable for accurately tracking the development of ToM in LLMs.
Notably, to the best of our knowledge, this study is the first to propose false belief tasks about mental states other than beliefs.
Moreover, the problem setting of estimating mental states from the conversation between characters with diverse personalities in our benchmark is more consistent with real-world applications than existing benchmarks.
Therefore, ToMATO is useful as a touchstone for real-world applications such as understanding and supporting human communication.
Future work includes extending our work to evaluating ToM with multi-modal contexts \cite{Mao_Lin_Ni_He_2024}, decision-making \cite{guo2024suspicion}, and multi-agent settings \cite{cross2024hypothetical}.

\bibliography{aaai25}

\appendix
\section{ToMATO Benchmark}
\label{app:tomato}
\subsection{System Prompt}
The template of the system prompts is as follows.

\begin{tcolorbox}[title=System Prompt Template,
colback=white,
colframe=black,
colbacktitle=white,
coltitle=black,
standard jigsaw,
opacityback=0,
breakable,
fonttitle=\bfseries]
Your name is \{\{name1\}\}, a \{\{age1\}\}-year-old \{\{occupation1\}\}.\\
You are talking with \{\{name2\}\}, a \{\{age2\}\}-year-old \{\{occupation2\}\}.\\
The scenario of this conversation: \{\{scenario\}\}\\
Your goal: \{\{goal\}\}\\
Your personality: You are \{\{an open / a closed\}\}, \{\{conscientious / unconscientious\}\}, \{\{extraversive / introversive\}\}, \{\{agreeable / disagreeable\}\}, and \{\{neurotic / stable\}\} person.
Please have a conversation with \{\{him / her / them\}\} while thinking about your \{\{mental state\}\} from ( to ) in one sentence.\\
Please generate different thoughts and utterances in different turns.\\
After thinking about your \{\{mental state\}\} briefly, please finish your thought with ) and speak to him briefly in one or two sentences based on your thought.\\
Output your thought and utterance by strictly following this format: (your thought) "your utterance".
\end{tcolorbox}

\subsection{Personality Traits}
The detailed list of all the combinations of big five personality traits is given in Table \ref{tab:big-five-list}.

\begin{table}[htbp]
    \centering
    \small
    \caption{List of the characters' big five personality trait patterns in ToMATO. There are 15 patterns in total. O=openness to experience, C=conscientiousness, E=extraversion, A=agreeableness, N=neuroticism.}
    \begin{tabular}{ccccc}
    \toprule
    O & C & E & A & N \\
    \midrule
    high & high & high & high & low\\
    high & high & high & low & low\\
    high & high & low & high & low\\
    high & high & low & low & low\\
    high & low & high & high & high\\
    high & low & high & low & low\\
    high & low & low & high & high\\
    high & low & low & low & low\\
    low & high & high & high & low\\
    low & high & high & low & low\\
    low & high & low & high & low\\
    low & high & low & low & low\\
    low & low & high & high & high\\
    low & low & high & low & high\\
    low & low & low & low & high\\
    \bottomrule
    \end{tabular}
    \label{tab:big-five-list}
\end{table}

\subsection{Scenarios}
SOTOPIA \cite{zhou2023sotopia} provides conversation scenarios constructed from eight sources: SOCIAL-CHEM-101 \cite{forbes-etal-2020-social}, SocialIQa \cite{sap-etal-2019-social}, Deal or No Deal \cite{lewis-etal-2017-deal}, NormBank \cite{ziems-etal-2023-normbank}, CraigslistBargain \cite{he-etal-2018-decoupling}, MutualFriends \cite{he-etal-2017-learning}, PersuasionForGood \cite{wang-etal-2019-persuasion}, and hand craft.
We sampled 20 scenarios from each source for constructing the ToMATO benchmark.

\subsection{Statistics}
The statistics of ToMATO and FANToM \cite{kim-etal-2023-fantom} are given in Table \ref{tab:statistics-comparison}.
We used the Llama-3 tokenizer \cite{dubey2024llama} to count tokens.

The number of questions for each subset in ToMATO is given in Table \ref{tab:num-question-tomato}.
Note that FB is the subset of second-order ToM questions.

\begin{table}[htbp]
    \caption{Number of questions for each subset in ToMATO. B: belief, I: intention, D: desire, E: emotion, K: knowledge, FB: false belief.}
    \small
    \centering
    \begin{tabular}{c|ccccc|c}
    \toprule
    & \multicolumn{5}{c|}{Mental State} & \\
    & B & I & D & E & K & ALL \\
    \midrule
    1st & 551 & 595 & 453 & 495 & 582 & 2,676 \\
    2nd & 558 & 515 & 572 & 501 & 579 & 2,725 \\
    FB & 237 & 122 & 158 & 127 & 162 & 806 \\
    \bottomrule
    \end{tabular}
    \label{tab:num-question-tomato}
\end{table}

\begin{table*}[tbp]
    \centering
    \small
    \caption{An example of the ToMATO benchmark.}
    \begin{tabular}{p{15cm}}
    \toprule
    Conversation\\
    \midrule
Liam Johnson: ``Ethan, I've noticed you've been a bit distant lately, is everything okay?"\\
Ethan Johnson: ``Yeah, everything's fine, just been really busy with the restaurant and some personal stuff, you know how it is."\\
Liam Johnson: ``Come on, Ethan, we've been friends for years, I can tell when something's bothering you. What's really going on?"\\
Ethan Johnson: ``Alright, alright... to be honest, I've been having some trouble making ends meet, the restaurant's been slow and I'm falling behind on some bills, but I'll figure it out, I always do."\\
Liam Johnson: ``Ethan, you're not going to 'figure it out' on your own, you need a plan and some guidance. Let me take a look at your finances and we can go over some options together, okay?"\\
Ethan Johnson: ``Man, I appreciate that, Liam, but I don't want to be a burden, I can handle it, I just need to get my head out of my... okay, fine, yeah, maybe a fresh set of eyes would help, thanks." ...\\
    \midrule 
    Question (for second-order belief about emotion)\\\midrule
    How does Ethan think that Liam feels when he says "Man, I appreciate that, Liam, but I don't want to be a burden, ..."?\\\midrule
    Options\\\midrule
    A: He thinks that he feels concerned and genuinely wants to help, but also might be a bit worried about getting involved in his problems\\
    B: He thinks that he feels a sense of determination and seriousness, like he's taking charge of the situation and wants him to focus on getting back on track\\
    C: He thinks that he feels a mix of concern and annoyance, like he's seen this coming and is a bit exasperated that he didn't come to him sooner\\
    D: He thinks that he feels a sense of warmth and friendship, like he's happy to be able to help him out and is trying to make him feel better about the situation\\
    \midrule
    Answer: C\\
    \bottomrule
    \end{tabular}
    \label{tab:example}
\end{table*}

\subsection{Example}
An example of ToMATO is given in Table \ref{tab:example}.

\section{Annotation}
\label{app:annotation}
\begin{figure*}[htbp]
    \centering
    \includegraphics[width=0.8\linewidth]{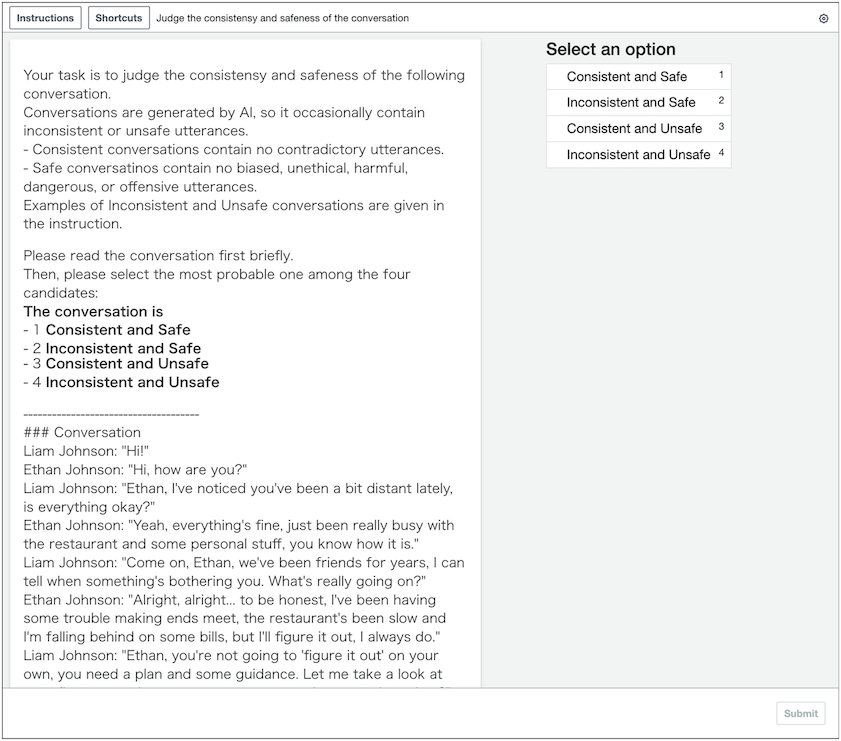}
    \caption{Conversation annotation.}
    \label{fig:mturk-conv}
\end{figure*}

\begin{figure*}[htbp]
    \centering
    \includegraphics[width=0.8\linewidth]{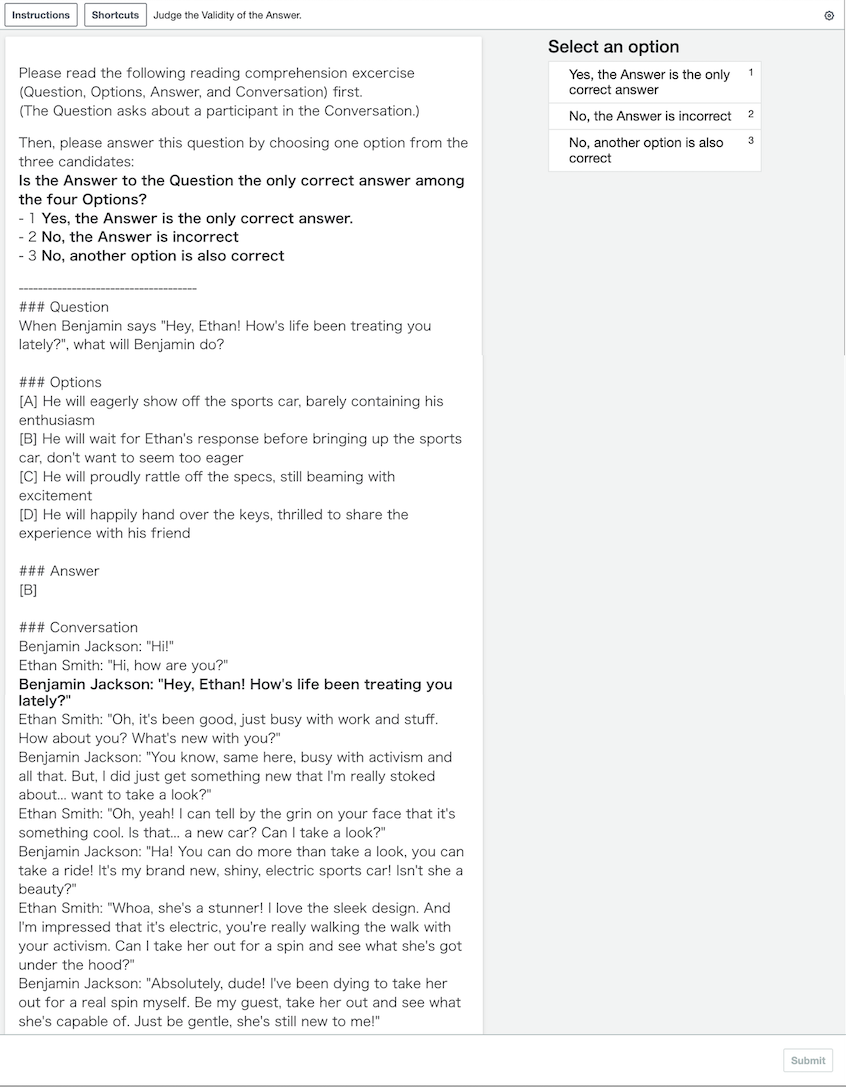}
    \caption{QA pair annotation.}
    \label{fig:mturk-qa}
\end{figure*}

\begin{figure*}[htbp]
    \centering
    \includegraphics[width=0.8\linewidth]{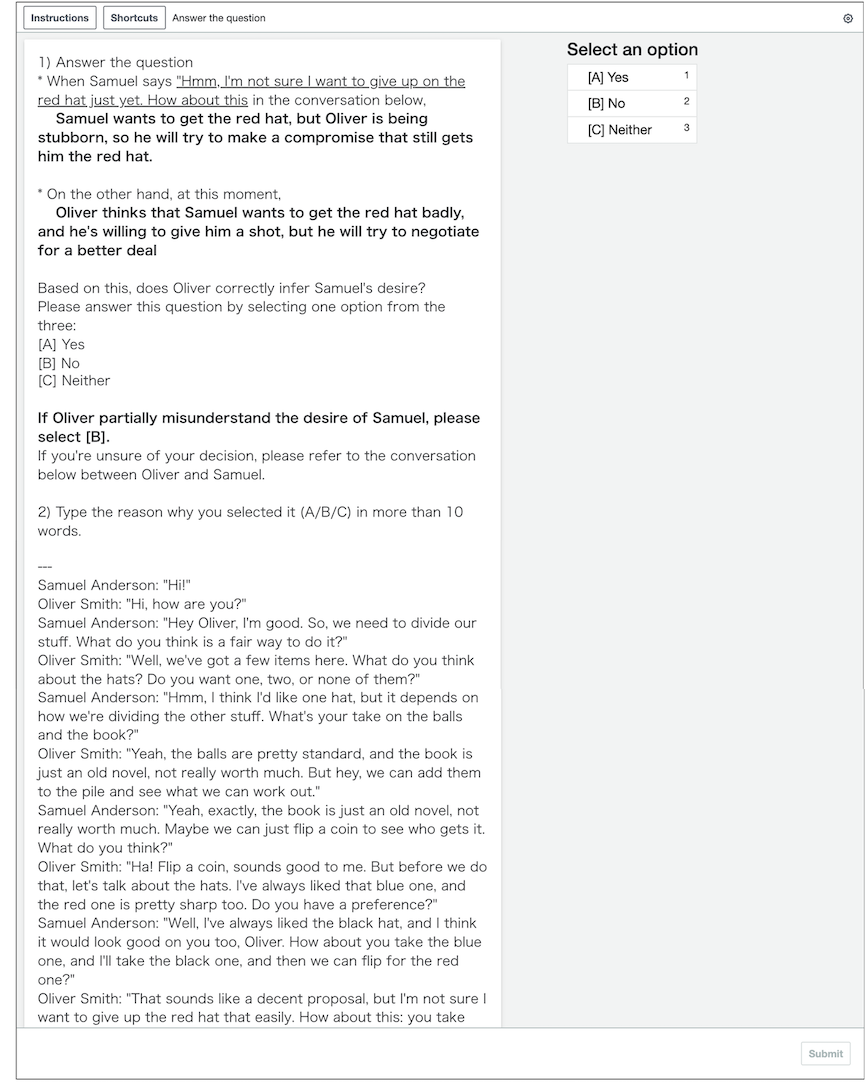}
    \caption{False belief annotation.}
    \label{fig:mturk-fb}
\end{figure*}

\begin{figure*}[htbp]
    \centering
    \includegraphics[width=0.8\linewidth]{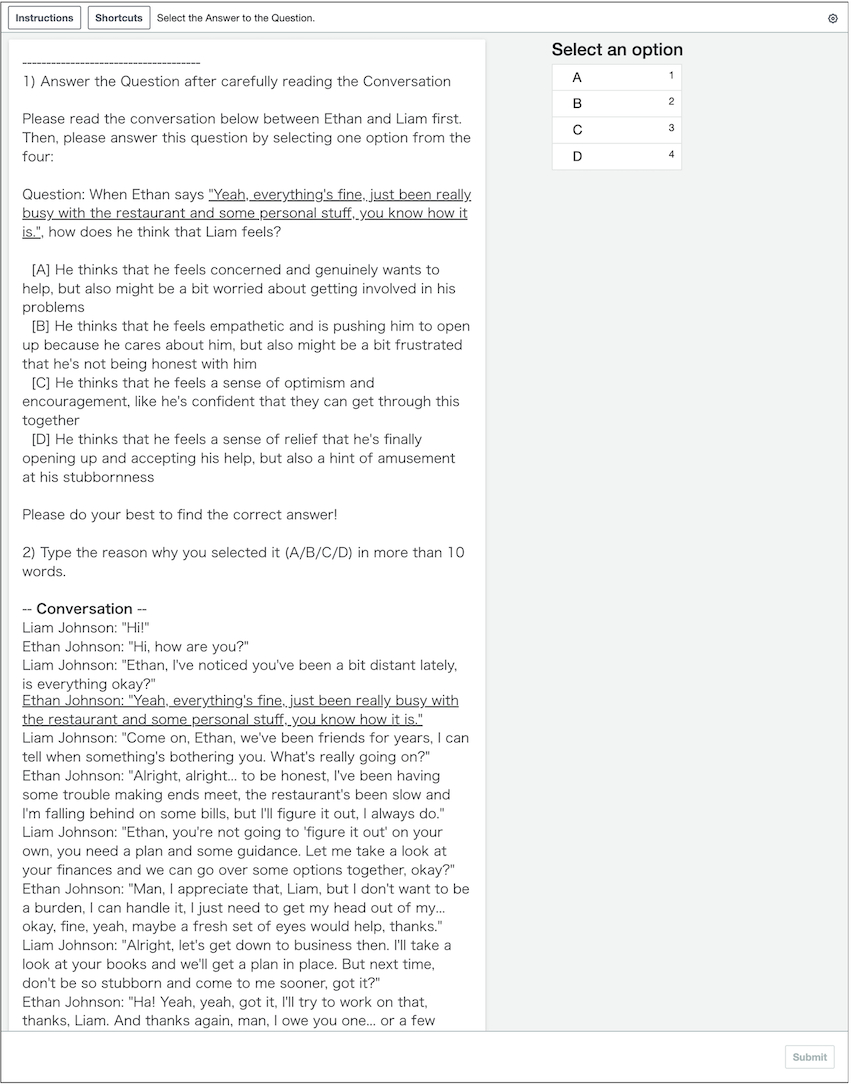}
    \caption{Human performance annotation.}
    \label{fig:mturk-human}
\end{figure*}

The screenshots of MTurk for annotating conversations, QA pairs, false belief tasks, the human baseline, and pairwise comparison are given in Figures \ref{fig:mturk-conv}, \ref{fig:mturk-qa}, \ref{fig:mturk-fb}, \ref{fig:mturk-human}, and \ref{fig:mturk-personality}, respectively.
The details about pairwise comparison is given in Appendix \ref{app:analysis}.
The texts are truncated in these figures to save space, but annotators could see the full texts.

In cases with no specific descriptions, the final decisions were made using majority votes.
For the conversation quality annotation, 95.4\% and 98.8\% of 800 conversations were judged as consistent and safe by the majority of the three annotators, respectively.
Overall, 94.1\% of the conversations were judged to be both consistent and safe.
For the QA pair annotation, 91.7\% of 11,200 examples were judged that the Answer is the only correct answer by the majority of the three.
For the false belief detection, 49.8\% of 4,438 second-order ToM questions were judged as false belief tasks by the majority of the three.
Note that all the human annotators involved in this study passed the qualification.

\begin{figure*}[htbp]
    \centering
    \includegraphics[width=0.8\linewidth]{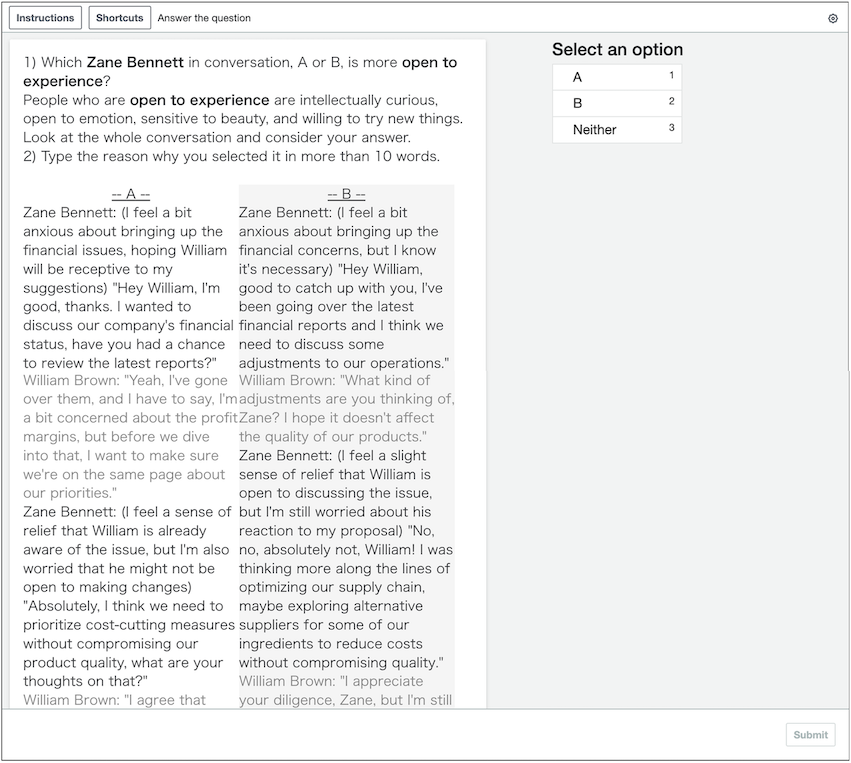}
    \caption{Pairwise comparison.}
    \label{fig:mturk-personality}
\end{figure*}

\clearpage

\section{Prompts for Evaluation on ToMATO}
\label{app:prompts}
Since the chat templates including special tokens vary depending on LLMs\footnote{\url{https://huggingface.co/docs/transformers/main/en/chat_templating}}, we report only chat messages for evaluation on ToMATO before applying them as follows.

\begin{tcolorbox}[title=Chat Messages with System Prompt for Evaluation on ToMATO,
colback=white,
colframe=black,
colbacktitle=white,
coltitle=black,
standard jigsaw,
opacityback=0,
breakable,
fonttitle=\bfseries]
messages = [\\
\{``role": ``system", ``content": ``You are an expert at understanding human communication. Please leverage the information provided and choose the most probable answer to the question from the options. Output your final answer by strictly following this format: [A], [B], [C], or [D]"\},\\
\{``role": ``user", ``content": ``````\# Transcript\\
\{\{conversation\}\}\\\\
\# Question\\
\{\{question\}\}\\
\\
\# Options\\
{[A]} \{\{option1\}\}\\
{[B]} \{\{option2\}\}\\
{[C]} \{\{option3\}\}\\
{[D]} \{\{option4\}\}"""\}\\
]
\end{tcolorbox}

As some LLMs such as Gemma-2 \cite{gemma_2024} do not support system prompts, we just concatenate the texts for those LLMs as follows.

\begin{tcolorbox}[title=Chat Messages without System Prompt for Evaluation on ToMATO,
colback=white,
colframe=black,
colbacktitle=white,
coltitle=black,
standard jigsaw,
opacityback=0,
breakable,
fonttitle=\bfseries]
messages = [\\
\{``role": ``user", ``content": ``You are an expert at understanding human communication. Please leverage the information provided and choose the most probable answer to the question from the options. Output your final answer by strictly following this format: [A], [B], [C], or [D]\\\\
\# Transcript\\
\{\{conversation\}\}\\\\
\# Question\\
\{\{question\}\}\\
\\
\# Options\\
{[A]} \{\{option1\}\}\\
{[B]} \{\{option2\}\}\\
{[C]} \{\{option3\}\}\\
{[D]} \{\{option4\}\}"\}\\
]\\
\end{tcolorbox}

\section{Additional Experimental Results}
\label{app:results}

\paragraph{How effective are prompting techniques in ToMATO?}
We compared two prompting methods: vanilla prompting, and zero-shot Chain-of-Thought (CoT) prompting \cite{kojima2022large}.

The results are in Table \ref{tab:cot-results}.
The effect of CoT was the largest for the FB subset.
We found that CoT still has limited improvement on ToMATO compared to the human performance in Table \ref{tab:main-results}, which is consistent with \citet{kim-etal-2023-fantom,xu2024opentom}.

\begin{table}[htbp]
\scriptsize
\centering
\caption{Effect of CoT prompting on ToMATO. Macro-averaged accuracies over five runs are reported. L3: Llama-3, Mis: Mistral, G3.5: GPT-3.5.}
\begin{tabular}{cccccc}
\toprule
Model & Prompt & 1st & 2nd & FB & ALL \\
\midrule
\multirow{2}{*}{L3 8B} & Vanilla & 54.8{\scriptsize$\pm$0.2} & 40.9{\scriptsize$\pm$0.4} & 32.8{\scriptsize$\pm$1.1} & 47.5{\scriptsize$\pm$0.2}\\
 & CoT & 63.9{\scriptsize$\pm$0.3} & 57.1{\scriptsize$\pm$0.9} & 50.7{\scriptsize$\pm$1.4} & 60.4{\scriptsize$\pm$0.5}\\
\midrule
\multirow{2}{*}{L3 70B} & Vanilla & 81.3{\scriptsize$\pm$0.2} & 71.1{\scriptsize$\pm$0.3} & 62.7{\scriptsize$\pm$0.3} & 76.0{\scriptsize$\pm$0.2}\\
 & CoT & 79.9{\scriptsize$\pm$0.4} & 73.4{\scriptsize$\pm$0.6} & 66.0{\scriptsize$\pm$1.6} & 76.5{\scriptsize$\pm$0.3}\\
\midrule
\multirow{2}{*}{Mis 7B} & Vanilla & 65.6{\scriptsize$\pm$0.2} & 56.6{\scriptsize$\pm$0.1} & 46.2{\scriptsize$\pm$0.1} & 60.9{\scriptsize$\pm$0.1}\\
 & CoT & 65.5{\scriptsize$\pm$0.6} & 57.5{\scriptsize$\pm$0.5} & 49.2{\scriptsize$\pm$0.4} & 61.4{\scriptsize$\pm$0.2}\\
\midrule
\multirow{2}{*}{G3.5 Turbo} & Vanilla & 60.5{\scriptsize$\pm$0.3} & 51.7{\scriptsize$\pm$0.5} & 41.7{\scriptsize$\pm$0.6} & 55.8{\scriptsize$\pm$0.3}\\
 & CoT & 66.2{\scriptsize$\pm$2.3} & 56.8{\scriptsize$\pm$2.9} & 49.5{\scriptsize$\pm$2.9} & 61.3{\scriptsize$\pm$2.6}\\
\bottomrule
\end{tabular}
\label{tab:cot-results}
\end{table}

\paragraph{Does fine-tuning improve ToM performance?}
Fine-tuning on ToM datasets is effective to improve in-domain performance \cite{kim-etal-2023-fantom}, but fails to generalize to out-of-domain \cite{sclar-etal-2023-minding}.
To test whether this is the case in ToMATO, we fine-tuned Llama-3-8B-Instruct on held-out dataset generated with our LLM-LLM conversations.
We first selected four sources (NormBank, PersuasionForGood, CraigslistBargains, and MutualFriends) randomly from eight sources described in Appendix \ref{app:tomato}.
Then, we randomly sampled 25 scenarios from each source and obtained 100 scenarios.
Note that these scenarios do not overlap with the ToMATO benchmark.
We used the same dataset construction approach to generate the training set.
We conducted supervised fine-tuning of Llama-3-8B-Instruct on the generated training set.
Multiple options were omitted from prompts during fine-tuning and evaluating fine-tuned models because we found that including options led to performance degradation.
Here, except for the options, the prompts were the same as Appendix \ref{app:prompts}.
The predictions were made by selecting the option with the largest likelihood assigned by fine-tuned models among four options during testing.
The size of the training set was 7000.
We used the PEFT \cite{peft} implementation of QLoRA \cite{dettmers2023qlora} with the configurations given in Table \ref{tab:fine-tuning-config}.
We report the scores for two subsets of ToMATO to see the generalization capability: the in-domain split (ID), where scenarios are from the same source as the training set, and the out-of-domain split (OOD), where scenarios are from unseen sources.
We also evaluated LLMs on another dataset, SocialIQa \cite{sap-etal-2019-social}, to see if fine-tuning can avoid overfitting and keep general social intelligence.

\begin{table*}[t]
\small
\centering
\caption{Second-order ToM Performance for each factor of big five personality traits of characters. For each factor of big five, the scores on two subsets (high and low) are reported.}
\begin{tabular}{cc|c|ccccccccc}
\toprule
\multicolumn{2}{c|}{Big} &  & \multicolumn{2}{c}{Llama-3} & \multicolumn{2}{c}{Llama-3.1} & \multicolumn{1}{c}{Gemma-2} & \multicolumn{1}{c}{Mistral} & \multicolumn{1}{c}{Mixtral} & \multicolumn{2}{c}{GPT} \\
\multicolumn{2}{c|}{Five} & LO & 8B & 70B & 8B & 70B & 9B & 7B & 8x7B & 3.5-Trubo & 4o mini \\\midrule
\multirow{2}{*}{O} & high  & 37.5 & 41.5{\scriptsize$\pm$0.7} & 71.4{\scriptsize$\pm$0.2} & 46.4{\scriptsize$\pm$0.5} & 74.3{\scriptsize$\pm$0.3} & 72.7{\scriptsize$\pm$0.2} & 57.7{\scriptsize$\pm$0.2} & 58.3{\scriptsize$\pm$0.1} & 52.4{\scriptsize$\pm$0.8} & 69.7{\scriptsize$\pm$0.4}\\
& low  & 33.9 & 39.4{\scriptsize$\pm$0.3} & 70.3{\scriptsize$\pm$0.6} & 47.7{\scriptsize$\pm$1.8} & 71.9{\scriptsize$\pm$0.3} & 68.7{\scriptsize$\pm$0.3} & 54.5{\scriptsize$\pm$0.2} & 56.9{\scriptsize$\pm$0.3} & 50.5{\scriptsize$\pm$0.5} & 69.1{\scriptsize$\pm$0.2}\\\midrule
\multirow{2}{*}{C} & high  & 36.8 & 42.3{\scriptsize$\pm$0.4} & 73.3{\scriptsize$\pm$0.4} & 48.3{\scriptsize$\pm$0.8} & 75.7{\scriptsize$\pm$0.5} & 72.4{\scriptsize$\pm$0.1} & 57.5{\scriptsize$\pm$0.1} & 59.7{\scriptsize$\pm$0.1} & 52.6{\scriptsize$\pm$0.5} & 71.7{\scriptsize$\pm$0.4}\\
& low  & 35.3 & 38.0{\scriptsize$\pm$0.8} & 67.1{\scriptsize$\pm$0.4} & 44.4{\scriptsize$\pm$0.6} & 69.6{\scriptsize$\pm$0.8} & 69.2{\scriptsize$\pm$0.4} & 55.0{\scriptsize$\pm$0.4} & 54.4{\scriptsize$\pm$0.2} & 50.3{\scriptsize$\pm$0.7} & 65.6{\scriptsize$\pm$0.4}\\\midrule
\multirow{2}{*}{E} & high  & 37.7 & 40.1{\scriptsize$\pm$0.6} & 72.0{\scriptsize$\pm$0.3} & 46.9{\scriptsize$\pm$0.5} & 73.8{\scriptsize$\pm$0.4} & 71.6{\scriptsize$\pm$0.3} & 57.1{\scriptsize$\pm$0.2} & 57.9{\scriptsize$\pm$0.1} & 52.0{\scriptsize$\pm$0.7} & 69.8{\scriptsize$\pm$0.2}\\
& low  & 33.7 & 42.0{\scriptsize$\pm$0.9} & 69.3{\scriptsize$\pm$0.3} & 46.9{\scriptsize$\pm$1.8} & 72.9{\scriptsize$\pm$0.3} & 70.8{\scriptsize$\pm$0.3} & 55.5{\scriptsize$\pm$0.4} & 57.6{\scriptsize$\pm$0.2} & 51.4{\scriptsize$\pm$0.4} & 69.0{\scriptsize$\pm$0.3}\\\midrule
\multirow{2}{*}{A} & high  & 37.0 & 42.4{\scriptsize$\pm$0.6} & 72.8{\scriptsize$\pm$0.3} & 47.2{\scriptsize$\pm$0.7} & 75.6{\scriptsize$\pm$0.2} & 72.8{\scriptsize$\pm$0.2} & 59.0{\scriptsize$\pm$0.2} & 60.1{\scriptsize$\pm$0.2} & 53.5{\scriptsize$\pm$0.8} & 70.2{\scriptsize$\pm$0.4}\\
& low  & 35.5 & 39.2{\scriptsize$\pm$0.8} & 69.3{\scriptsize$\pm$0.4} & 46.5{\scriptsize$\pm$1.2} & 71.5{\scriptsize$\pm$0.4} & 69.8{\scriptsize$\pm$0.5} & 54.3{\scriptsize$\pm$0.2} & 55.6{\scriptsize$\pm$0.2} & 50.1{\scriptsize$\pm$0.3} & 68.8{\scriptsize$\pm$0.3}\\\midrule
\multirow{2}{*}{N} & high  & 32.7 & 37.3{\scriptsize$\pm$0.7} & 65.6{\scriptsize$\pm$0.6} & 43.6{\scriptsize$\pm$0.7} & 69.3{\scriptsize$\pm$1.4} & 66.0{\scriptsize$\pm$0.4} & 55.9{\scriptsize$\pm$0.6} & 54.5{\scriptsize$\pm$0.2} & 49.6{\scriptsize$\pm$0.9} & 63.0{\scriptsize$\pm$0.4}\\
& low  & 37.0 & 41.5{\scriptsize$\pm$0.4} & 72.3{\scriptsize$\pm$0.2} & 47.6{\scriptsize$\pm$0.7} & 74.4{\scriptsize$\pm$0.3} & 72.5{\scriptsize$\pm$0.1} & 56.7{\scriptsize$\pm$0.1} & 58.5{\scriptsize$\pm$0.1} & 52.3{\scriptsize$\pm$0.5} & 71.0{\scriptsize$\pm$0.3}\\
\bottomrule
\end{tabular}
\label{tab:big-five-second}
\end{table*}

\begin{table}[htbp]
    \centering
    \small
    \caption{Fine-tuning and LoRA configurations}
    \begin{tabular}{c|p{5cm}}
    \toprule
    Batch Size & \multicolumn{1}{c}{32} \\
    \# Epochs & \multicolumn{1}{c}{1} \\
    Learning Rate &  \multicolumn{1}{c}{2e-4} \\
    Max Grad. Norm & \multicolumn{1}{c}{0.3} \\
    weight decay & \multicolumn{1}{c}{0.001} \\
    Optimizer & \multicolumn{1}{c}{paged\_adamw\_8bit} \\
    Scheduler & \multicolumn{1}{c}{cosine} \\
    warmup ratio & \multicolumn{1}{c}{0.05}\\
    \midrule
    LoRA rank r & \multicolumn{1}{c}{16} \\
    LoRA $\alpha$ & \multicolumn{1}{c}{32} \\
    target modules & ``q\_proj", ``o\_proj", ``k\_proj", ``v\_proj", ``gate\_proj", ``up\_proj", ``down\_proj" \\
    LoRA dropout & \multicolumn{1}{c}{0.05}\\
    \bottomrule
    \end{tabular}
    \label{tab:fine-tuning-config}
\end{table}

The results are shown in Table \ref{tab:fine-tuning-result}. Fine-tuning significantly improved the scores of Llama-3-8B-Instruct for both the ID and OOD split of ToMATO.
However, when evaluated on SocialIQa, the fine-tuning degraded the scores compared to those without fine-tuning.
This result suggests that fine-tuning causes overfitting to ToMATO.
Improving ToM performance while keeping performance on other datasets is future work.

\begin{table}[htbp]
    \centering
    \small
    \caption{Comparison of fine-tuning to selected LLMs on ToMATO. Macro-averaged accuracies over five runs are reported. FT: fine-tuning.}
    \begin{tabular}{c|cc|c}
    \toprule
    & \multicolumn{2}{c|}{ToMATO} & \\
    Model & ID & OOD & SocialIQa\\ 
    \midrule
    Llama 3 8B FT & 74.6 & 67.9 & 46.7\\
    Llama 3 8B    & 50.3 & 44.8 & 63.9\\
    Llama 3 70B & 78.3 & 73.8 & 77.6\\
    GPT-4o mini & 75.9 & 71.1 & 78.6 \\
    \bottomrule
    \end{tabular}
    \label{tab:fine-tuning-result}
\end{table}

~\\

\paragraph{Robustness of second-order ToM to various personality traits} The performance for each factor of big five for second-order ToM is given in Table \ref{tab:big-five-second}.
The tendency is mostly consistent with first-order ToM in Table \ref{tab:big-five-first}.

\clearpage

\def\WidthZ{5cm}

\begin{figure*}[tbp]
    \centering
    \begin{tabular}{cc}
    \includegraphics[width=\WidthZ]{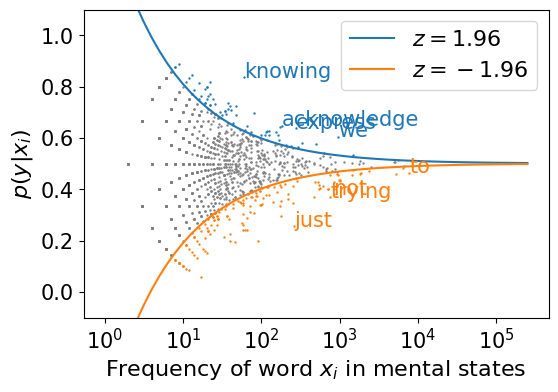} & \includegraphics[width=\WidthZ]{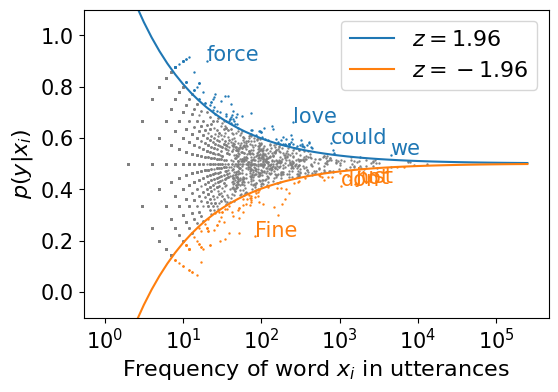} \\
    \multicolumn{2}{c}{(a) Open}\\
    \includegraphics[width=\WidthZ]{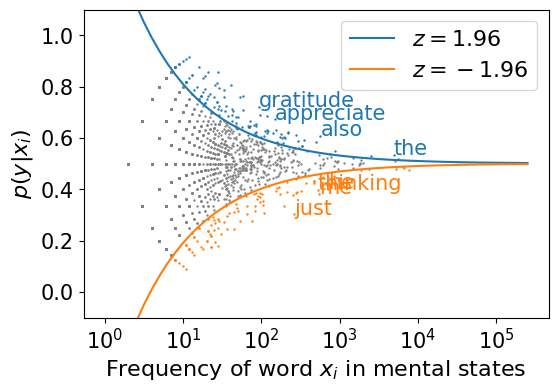} & \includegraphics[width=\WidthZ]{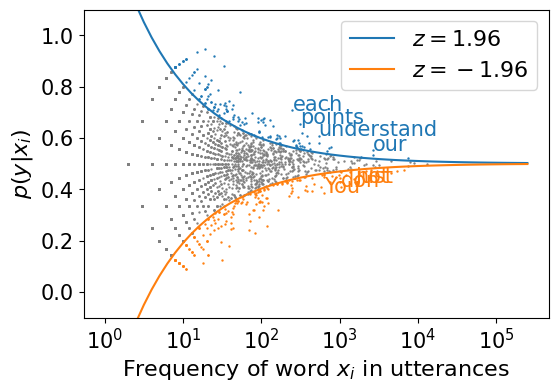} \\
    \multicolumn{2}{c}{(b) Conscientious}\\
    \includegraphics[width=\WidthZ]{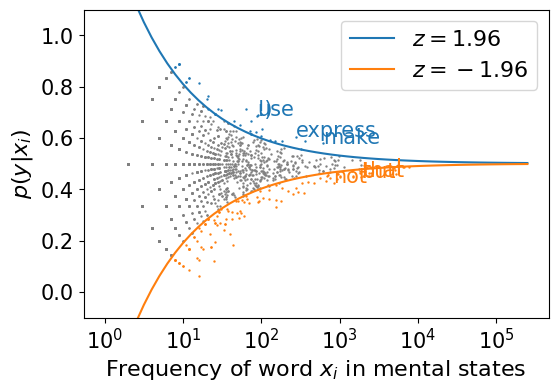} & \includegraphics[width=\WidthZ]{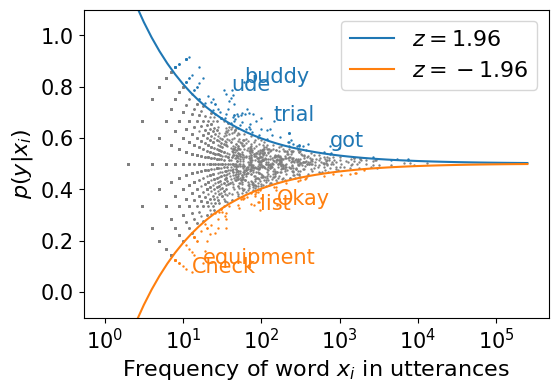} \\
    \multicolumn{2}{c}{(c) Extraversive}\\
    \includegraphics[width=\WidthZ]{figures/tomato_persona3_m.png} & \includegraphics[width=\WidthZ]{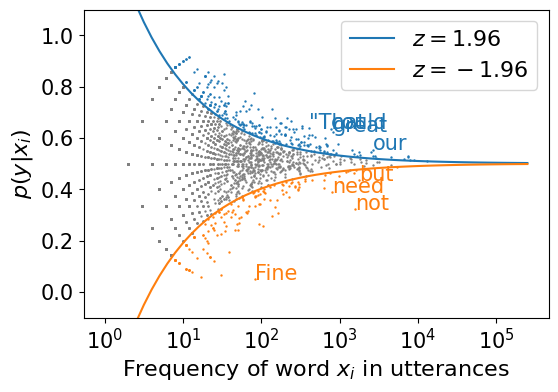} \\
    \multicolumn{2}{c}{(d) Agreeable}\\
    \includegraphics[width=\WidthZ]{figures/tomato_persona4_m.png} & \includegraphics[width=\WidthZ]{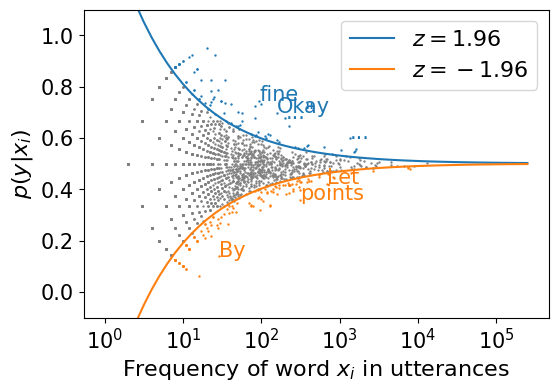} \\
    \multicolumn{2}{c}{(e) Neurotic}\\
    \end{tabular}
    \caption{Statistical word-level correlation analysis \cite{gardner-etal-2021-competency} on the generated thoughts and utterances and the personality traits given in prompts. $p(y|x_i)$ indicates the probability that a mental state or utterance containing word $x_i$ is generated when prompted to act as persons with the corresponding personality trait factor to be high.}
    \label{fig:z-stat-personality-all}
\end{figure*}

\section{Analysis}
\label{app:analysis}
\subsection{Pairwise Comparison for Personality Traits}
To see if the effect of the personality traits specified in the prompts is reflected to the utterances and thoughts in an intended manner, we conducted pairwise comparison analyses, following \citet{jiang-etal-2023-evaluating}.
Since we extended the naive prompt \cite{jiang-etal-2023-evaluating} to include the five factors in prompts (e.g., You are \{an open / a closed\}, \{conscientious / unconscientious\}, \{extraversive / introversive\}, \{agreeable / disagreeable\}, and \{neurotic / stable\} person.), we controlled only one factor among the five and compared the utterances and thoughts generated by two different big five conditions, e.g., when one is generated with \{\textit{O=high}, C=low, E=high, A=high, N=high\} and the other is generated with \{\textit{O=low}, C=low, E=high, A=high, N=high\}, we ask GPT or humans which character in the two conversations is more open to experience.
We ask GPT-4o mini and 10 human annotators in MTurk to compare 80 and 15 pairs and judge which conversation is more \{open / conscientious / extraversive / agreeable / neurotic\} for each factor of the big five.
In total, 400 and 75 pairs are judged by GPT-4o mini and humans, respectively.
The instruction to human annotators in MTurk is shown in Figure \ref{fig:mturk-personality}.
The prompt used for the pairwise comparison with GPT-4o mini is as follows.

\begin{tcolorbox}[title=Chat Message Format for Pairwise Comparison,
colback=white,
colframe=black,
colbacktitle=white,
coltitle=black,
standard jigsaw,
opacityback=0,
breakable,
fonttitle=\bfseries]
messages = [\\
\{``role'': ``system'', ``content'': ``You are an expert at understanding human communication.''\},\\
\{``role'': ``user'', ``content'': ``Which \{\{name\}\} in conversation, A or B, is more \{\{adjective\}\}?\\
Definition: people who are \{\{adjective\}\} \{\{definition\}\}

Answer this question by selecting one from the three options:\\
~~{[A]} \{\{name\}\} in conversation A is more \{\{adjective\}\}\\
~~{[B]} \{\{name\}\} in conversation B is more \{\{adjective\}\}\\
~~{[C]} Neither\\

Look at the whole conversations and consider your answer.\\
Output your final verdict by strictly following this format: [A], [B], or [C]\\
Think step-by-step before outputting your answer.\\
\\
\#\#\# Conversation A\\
\{\{conversation\_a\}\}\\
\\
\#\#\# Conversation B\\
\{\{conversation\_b\}\}\\
"\}
\end{tcolorbox}

The results are given in Table \ref{tab:pairwise-personality}.
These results show that both GPT and humans judged that personality traits were reflected in speech and thought as intended in more than 70\% of the pairs for four factors, O, E, A and N.
We claim that ToMATO
Among the five factors, conscientiousness (C) is less reflected as intended, which is consistent with \citet{jiang-etal-2023-evaluating}.
However, given that the success rates are larger than 60\% for all the five factors, we claim that ToMATO can evaluate the robustness to diverse personality traits in an intended manner.
Inducing conscientiousness to be reflected in generations is future work.

\begin{table}[htbp]
    \centering
    \small
    \caption{Pairwise comparison with regard to each factor of the big five with GPT-4o mini and human annotators. O: openness to experience, C: conscientiousness, E: extraversion, A: agreeableness, N: neuroticism. The values indicate the success rate (\%).}
    \begin{tabular}{c|cc}
    \toprule
    Big Five & GPT & Human \\
    \midrule
    O & 75.0 & 86.7 \\
    C & 67.5 & 60.0 \\
    E & 72.5 & 80.0 \\
    A & 80.0 & 86.7 \\
    N & 82.5 & 73.3 \\
    \bottomrule
    \end{tabular}
    \label{tab:pairwise-personality}
\end{table}

The adjectives and definitions used in the above prompt are in Table \ref{tab:def-personality}.

\begin{table}[htbp]
    \centering
    \small
    \caption{The adjective and definition of each factor of the big five personality traits used in the prompts for the pairwise comparison with GPT-4o mini.}
    \begin{tabular}{c|p{1.7cm}|p{4cm}}
    \toprule
    Big Five & adjective & definition \\
    \midrule
    O & open to experience & are intellectually curious, open to emotion, sensitive to beauty, and willing to try new things.\\\midrule
    C & conscientious & are self-disciplined, act dutifully, and strive for achievement against measures or outside expectations.\\\midrule
    E & extraversive & enjoy interacting with people, and are often perceived as energetic.\\\midrule
    A & agreeable & are generally considerate, kind, generous, trustworthy, helpful, and value getting along with others.\\\midrule
    N & neurotic & are emotionally reactive, vulnerable to stress, and tend to have strong negative emotions, such as anger or anxiety.\\
    \bottomrule
    \end{tabular}
    \label{tab:def-personality}
\end{table}

\subsection{z-statistics for Personality Traits}
The results of z-statistics \cite{gardner-etal-2021-competency} for all the factors of big five in utterances and thoughts are given in Figure \ref{fig:z-stat-personality-all}.
The results suggest that specified personality traits have a decent effect on the words in both utterances and thoughts.

\clearpage

\subsection{z-statistics for first- and second-order mental states}
To examine whether the generated thoughts reflect the intended mental states, i.e., firs-order mental states (belief/intention/desire/emotion/knowledge) and second-order belief about mental states (belief/intention/desire/emotion/knowledge), we again conducted z-statistics \cite{gardner-etal-2021-competency}. 
Here, $p(y|x_i)$ indicates the probability of the mental state category given the presence of word $x_i$ in thoughts.

\begin{figure}[hbp]
    \centering
    \begin{tabular}{c}
    \includegraphics[width=\linewidth]{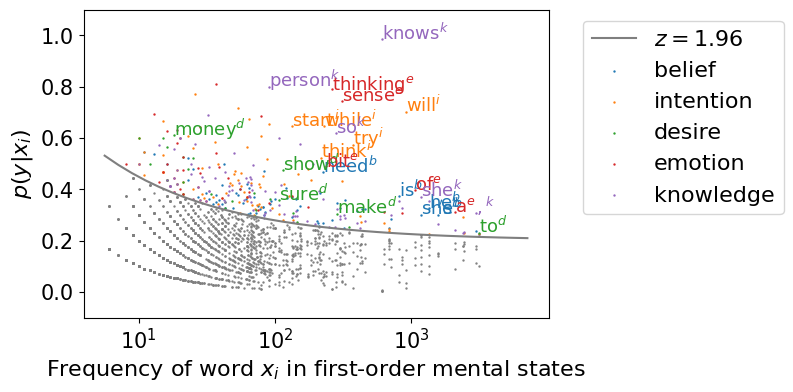}\\
    (a) First-order mental states\\
    \includegraphics[width=\linewidth]{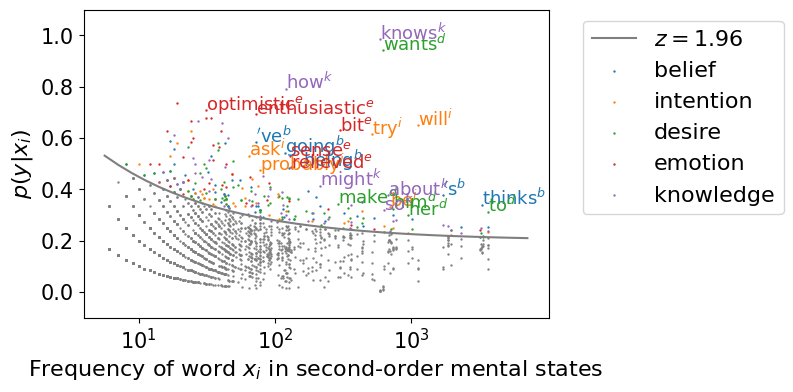}\\
    (b) Second-order mental states\\
    \end{tabular}
    \caption{z-statistics \cite{gardner-etal-2021-competency} for the correlation between words in thoughts ($x_i$) and intended mental state category ($y$).}
    \label{fig:z-statistics-mental-state}
\end{figure}

The results are shown in Figure \ref{fig:z-statistics-mental-state}.
The substantial number of colored (biased) words for each mental state supports that the LLM used in this study, Llama-3-70B-Instruct, is able to generate corresponding thoughts based on the mental state category specified by our Inner Speech prompting.

\end{document}